%% file: main_comm_physics.tex
\documentclass[11pt]{article}
\usepackage[utf8]{inputenc}
\usepackage[]{framed}
\input packages.tex

\usepackage[ruled,vlined]{algorithm2e}
\usepackage{nameref}

\pgfplotsset{compat=1.14}

\usepackage{overpic}

\usepackage{amsthm}
\newtheorem{lemma}{Lemma}
\newtheorem{theorem}{Theorem}
\newtheorem{corollary}{Corollary}
\theoremstyle{definition}
\newtheorem*{proof*}{Proof}

\newcommand{\aditione}[1]{}
\newcommand{\michaelone}[1]{}

\newcommand\eref{Eq.~\ref}
\newcommand\fref{Fig.~\ref}

\usepackage{hyperref}
\usepackage[font={small,it}]{caption}

\hypersetup{ %
	pdftitle={},
	pdfauthor={},
	pdfkeywords={},
	pdfborder=0 0 0,
	pdfpagemode=UseNone,
	colorlinks=true,
	linkcolor=magenta,
	citecolor=magenta,
	filecolor=magenta,
	urlcolor=magenta,
}

\usepackage{parskip}

\title{Learning Continuous Models for Continuous Physics}

\author{\normalsize
Aditi S. Krishnapriyan\footnote{Corresponding author: Aditi S. Krishnapriyan (aditik1@berkeley.edu)} \footnote{ASK and AFQ contributed equally} $^{1,2}$, Alejandro F. Queiruga$^{\dagger 3}$, \\ \vspace{-1ex} N. Benjamin Erichson$^{1,4}$, Michael W. Mahoney$^{1,2,4}$ \\
\vspace{3ex}
\small 
$^1$Lawrence Berkeley National Laboratory, 
$^2$University of California, Berkeley, \\
\small 
$^3$Google Research, 
$^4$International Computer Science Institute
}
\date{}
\begin{document}

\maketitle

\setcounter{page}{1}

\vspace{-0.3cm}
\input{_s0_abstract}
\input{_s1_introduction_comm_physics}

\input{_s4_results_comm_physics}

\input{_s5_conclusions_comm_physics}

\input{_s3_problem_overview_comm_physics}

\input{_s6_acknowledgements}

{
\bibliographystyle{unsrt}
\small

}

\input{_s7_supplementary}

\end{document}

%% file: packages.tex
\usepackage{array, makecell}
\usepackage{amsmath}
\usepackage{fullpage}
\usepackage{graphicx}
\usepackage{tabularx}           
\usepackage{float}              
\usepackage{comment}
\usepackage{multirow}
\usepackage{pgfplots}           
\usepackage{subfig}
\usepackage{amsmath,amssymb,amsfonts,amsbsy,amsfonts,latexsym, amsthm}
\usepackage{url}
\usepackage{authblk}
\newtheorem{definition}{Definition}
\newtheorem{remark}{Remark}

\usepackage{amsmath}

%% file: _s0_abstract.tex
\begin{abstract} 
Dynamical systems that evolve \emph{continuously} over time are ubiquitous throughout science and engineering. Machine learning (ML) provides data-driven approaches to model and predict the dynamics of such systems.
A core issue with this approach is that ML models are typically trained on discrete data, using ML methodologies that are not aware of underlying continuity properties.
This results in models that often do not capture any underlying continuous dynamics---either of the system of interest, or indeed of any related system. 
To address this challenge, we develop a convergence test based on numerical analysis theory. 
Our test verifies whether a model has learned a function that accurately approximates an underlying continuous dynamics.
Models that fail this test fail to capture relevant dynamics, rendering them of limited utility for many scientific prediction tasks; while models that pass this test enable both better interpolation and better extrapolation in multiple ways.
Our results illustrate how principled numerical analysis methods can be coupled with existing ML training/testing methodologies to validate models for science and engineering~applications.

\end{abstract}

%% file: _s1_introduction_comm_physics.tex
\section*{Introduction}

Dynamical systems---systems whose state varies over time---describe many chemical, physical, and biological processes. 
Thus, understanding and describing these dynamical systems is important for many scientific and engineering applications. 
Dynamical systems can often be described by \emph{differential} equations which evolve continuously in time, meaning that the domain of the solution spans a continuum~\cite{robinson2012introduction}. 
In such systems, the gap between any two timesteps can be subdivided into an infinite number of infinitely smaller timesteps. 
In practice, these systems are often identified via a finite set of discrete observational data, and 
%
there is a long history within scientific computing for dealing with this discrete-to-continuous gap: 
experimentally measuring scientific data at sufficiently fine timescales to resolve approximately-continuous dynamics of interest; formulating theory within function spaces of sufficient smoothness to guarantee certain continuity requirements; and 
developing numerical algorithms that come with appropriate stability and convergence guarantees.

Machine learning (ML) techniques have recently been shown to provide a powerful approach to model and learn from discrete data, and many scientific fields make extensive use of data-driven methods for describing~\cite{brunton2019data, calinon2012statistical, peters2007machine}, discovering~\cite{brunton2016discovering,raissi2018multistep,keller2021discovery}, identifying~\cite{rudy2019data,jin2020sympnets}, predicting~\cite{lutter2018deep,chen2019symplectic,erichson2020lipschitz,rusch2021long, wang2021bridging,lim2021noisy, jiahao2021knowledge, negiar2023learning}, and controlling~\cite{brunton2019data,morton2019deep, lambert2020objective,Li2020Learning, bachnas2014review} dynamics.
These approaches (see~\cite{karniadakis2021physics} for a survey) include purely data-driven methods that learn from observational data points~\cite{manojlovic2020applications}, adding constraints to ML methods that aim to respect the relevant physics~\cite{krishnapriyan2021characterizing, negiar2023learning}, and/or hybrid methods combining classical numerical solvers with (say) deep learning~\cite{kovachki2021neural, bar2019learning, pestourie2021physics}.

In many scientific and engineering applications, we observe measurements that yield a series of discrete data points $\left\{x_0, \, x_1, \, x_2, \, \dots, \, x_N\right\}$, where each point is spaced apart by some timestep size $\Delta t$.
There are many techniques from ML and statistical data analysis to learn data-driven input-output mappings ($G: x_n \rightarrow x_{n+1}$) that can provide an approximation for the next discrete timestep.
One popular class of data-driven input-output mappings is given by neural networks (NNs). A NN, denoted as $\mathcal{N}$, can be trained to predict  $x_{n+1}$ from $x_{n}$ by learning model parameters $\theta$:
\begin{align}
	x_{n+1} &= \mathcal{N}(x_n; \theta) .
	\label{eq:nn_approach}
\end{align} 

However, when considering continuous dynamical systems, there are challenges with this approach. 
Most obviously, this approach does not learn a continuous function~\cite{erichson2019physics,otto2019linearly,azencot2020forecasting,dubois2020data}; it simply learns a function that predicts subsequent discrete time steps.
This is to be expected, as this model is optimized to make (discrete) point estimates, i.e., to predict solutions at specific (discrete) points. 
For this reason, predicting future states of a dynamical system with this approach can result in compounding errors of the dynamics over time~\cite{asadi2019combating, lambert2020objective}.

A related approach is to assume that the discrete data points can be modeled and described by a continuous differential equation of the form,
\begin{equation}
	\frac{\mathrm{d}x(t)}{\mathrm{d}t} = F(x(t)),
	\label{eq:de}
\end{equation}
where $F$ is a function that describes the vector field.  
In some cases, there is an underlying true $F$, while in other cases it is simply a modeling assumption. A challenge is that we cannot derive $F$ from first-principles in many situations. 
Instead, we can use a data-driven approach for modeling $F$. For instance, an arbitrary NN architecture $\mathcal{N}$ can be used to model the vector field $F$,
\begin{equation}
	\frac{\mathrm{d}x(t)}{\mathrm{d}t} =  \mathcal{N}(x(t); \theta).
	\label{eq:node}
\end{equation}
This approach, so-called Neural Ordinary Differential Equations (ODE-Nets), has been proposed to model temporal systems~\cite{chen2018neural, ruthotto2020deep, queiruga2021stateful, massaroli2020dissecting, zhang2019anodev2, weinan2017proposal, rubanova2019latent}.

It is often assumed or simply taken for granted that ODE-Nets and other ML methods for ODEs automatically capture some continuous dynamics, either of the system that generated the data or of some related system~\cite{greydanus2019hamiltonian, du2020, greydanus2021piecewise, chen2020learning, jia2019neural}. 

However, due to how ODE-Nets are trained, i.e., to predict solutions at specific (discrete) points, these models can easily fail to learn even the simplest continuous dynamical systems~\cite{queiruga2020continuous, ott2020neural}, even when they accurately fit the temporal discretization (i.e., the discrete training points and testing points).
An ODE-Net that incorrectly learns a continuous model will simply provide high-quality discrete time predictions --- i.e., it is not a ContinuousNet but is simply a very good DiscreteNet~\cite{queiruga2020continuous}. 

Such a model will fail to extrapolate to new data points outside the temporal discretization, and it will fail to interpolate the solution at timesteps in between the discrete training data. 
It can also fail to correctly identify qualitative long-term behavior such as bifurcations~\cite{ker_disc_cont_92}.
As we demonstrate later, this Discrete-versus-Continuous distinction affects non-NN ML methods as well, even when they accurately fit the temporal discretization of the data.

\begin{figure*}[!t]
	\centering
	\subfloat[Training data (training trajectory)] {{ 
			\includegraphics[width=0.45\linewidth]{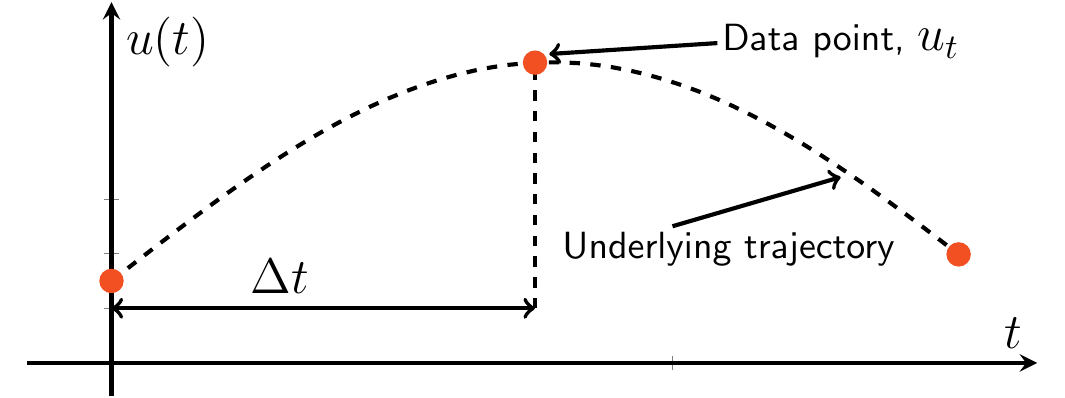}\label{subfig:discrete_training_data}
	}}
	\subfloat[Validation data (validation trajectory)] {{ \includegraphics[width=0.45\linewidth]{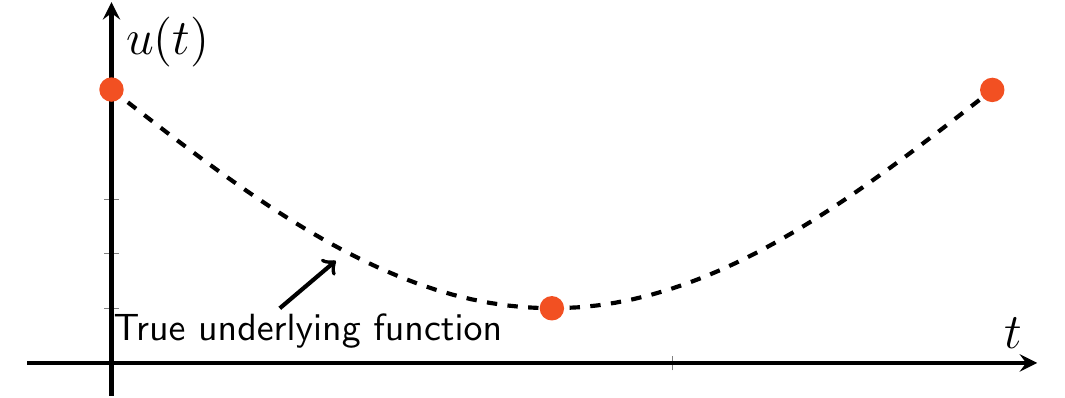}\label{subfig:discrete_validation_data}
	}}
	
	\subfloat[Discrete-only model] {{ 
			\includegraphics[width=0.45\linewidth]{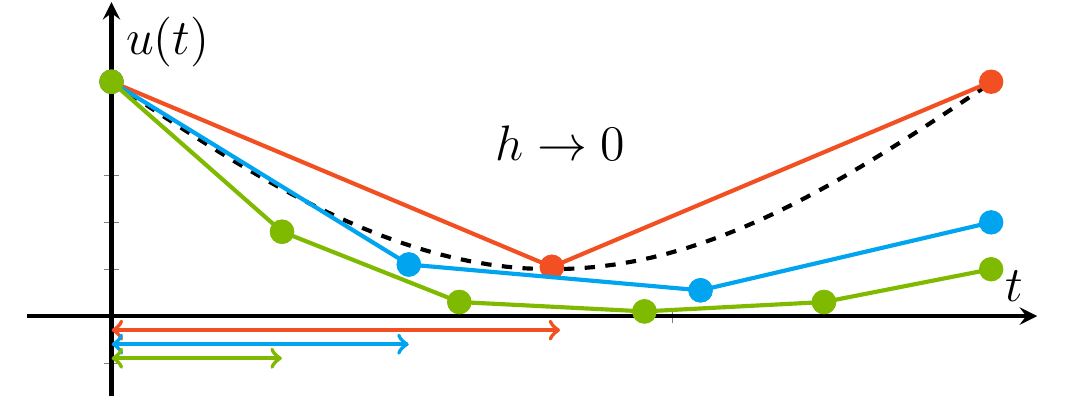} \label{subfig:discrete_only_model}
	}}
	\subfloat[Continuous model] {{ \includegraphics[width=0.45\linewidth]{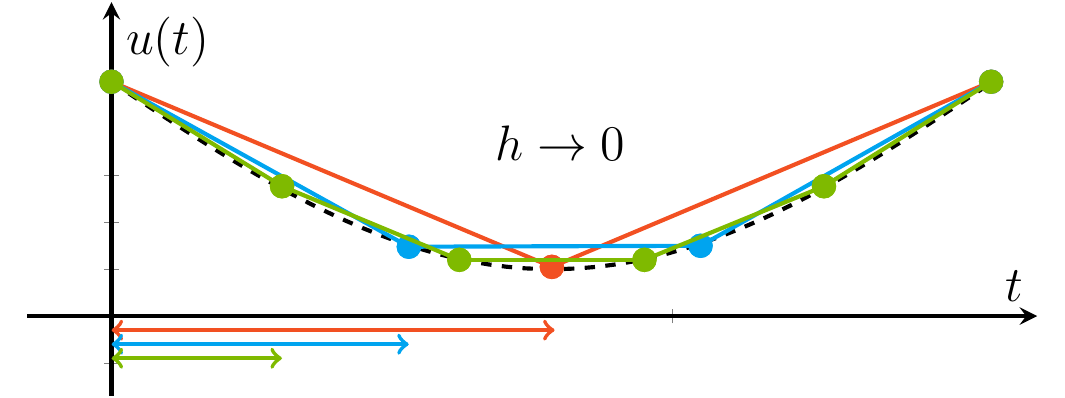} \label{subfig:continuous_model}
	}}
	
	\caption{\textbf{Learning to predict discrete points versus learning continuous dynamics.} 
		\protect\subref{subfig:discrete_training_data}
		A NN model learns from a discrete set of points (red dots) that are generated by a dynamical system with an underlying continuous trajectory (dashed black line). 
		\protect\subref{subfig:discrete_validation_data} After training, the model should be able to predict future data points that are lying on the same or a different trajectory, including data points that are irregularly sampled.
		\protect\subref{subfig:discrete_only_model}
		A model that has only learned to predict the discrete data points might accurately fit future points on the same trajectory sampled at rate $\Delta t$ (shown in red), but fail to predict points sampled at different rates. The blue and green lines evaluated with $h < \Delta t$ fall off of the underlying continuous trajectory.
		\protect\subref{subfig:continuous_model}
		A model that has learned an underlying continuous dynamics can converge to a continuous solution as $h \rightarrow 0$. 
		In this case, the blue and green lines evaluated with $h < \Delta t$ get closer to the underlying continuous trajectory.
	}
	\label{fig:schematic_overview}
\end{figure*}

Figure~\ref{fig:schematic_overview} illustrates the difference between a model that has learned to predict discrete data points and a model that has learned an underlying continuous dynamics. 
After training a model at a given discretization $\Delta t$, the trained model can be used to predict trajectories at arbitrary timestep sizes $h$ during inference. For validation, an error metric $Error(h)$ can be defined over a holdout trajectory that allows for evaluation with discretizations that are different than the data spacing.
Learning a discrete-only model (a DiscreteNet) means that only the discrete training points---and potentially testing points at the same discretization (i.e., when $h = \Delta t$)---are learned.
When evaluating using $Error(h=\Delta t)$, the model will appear to perform well.
The model may perform well on testing points with a similar discretization, but it will perform poorly for points sampled with other discretizations; that is, $Error(h\neq\Delta t)$ can be much worse than a discrete-only testing methodology would determine.
This will even occur when the discretization $h \rightarrow 0$, counter to expectations.
In contrast, learning a meaningfully continuous model (a ContinuousNet) means that the model can converge to a smooth solution as the discretization $h \rightarrow 0$, or at least that its error will decrease gradually and level off as $h \rightarrow 0$.  
(This will be true regardless of whether the learned continuous model corresponds to the true underlying model, even assuming that such a true model exists and/or is well-conditioned.)
In this case, the model will perform well for a broader range of temporal discretizations and thus have a better approximation of the continuous dynamical system.

In this work, we adapt methods from numerical analysis theory to develop a methodology to verify whether an ML model has learned a meaningfully continuous function that describes a dynamical system of interest. 
%
%
Specifically, we introduce a modified convergence test to verify and validate whether a model has learned continuous dynamics for a physical system. 
Our method allows us to verify that a model approximates a continuous differential operator, rather than only learning discrete points at a given temporal discretization, in the same sense that discrete algorithms from numerical analysis can be said to approximate continuous functions.
We also introduce the notion of a ContinuousNet to refer to an ODE-Net model that exhibits the convergence properties that are expected for a continuous time system:
\begin{definition}[ContinuousNet]
\label{def:continuousnet}
An ODE-Net,
\begin{equation*}
	\frac{d x(t)}{d t} = \mathcal{N}_\theta (x(t)), \quad x(0)=x_0 \in \mathbb{R}^n, \,\, \text{and} \,\,\, t\in\mathbb{R},
\end{equation*}
trained with a numerical integration scheme is a \emph{ContinuousNet} if it is convergent to a similar error as the error obtained by using the original training time step,
\begin{equation}
    \lim_{h\rightarrow 0}\mathrm{Error}(h) \lesssim \mathrm{Error}(\Delta t).    \label{eq:convergent_criterion}
\end{equation}
\end{definition}
This ContinuousNet convergence criteria is very similar to that of numerical analysis, whereby convergence is judged as $h\rightarrow0$, and can be evaluated with a similar methodology adapted for the ML setting.
The criteria of \eref{eq:convergent_criterion} represents a heuristic that takes into account the error from the learning process that can be observed in the error on the discrete-only validation task, $Error(\Delta t)$.
(Note that we are not claiming that this method will guarantee that we have learned the true solution---that would require additional assumptions, well-known in scientific computing---simply that we have learned some underlying continuous model of the data.)

To illustrate the utility of our approach, we demonstrate how meaningfully continuous models that pass our convergence test enable both better interpolation and better extrapolation in multiple ways. 
We show that such models can resolve fine-scale features of the solution, despite being trained only on coarse data, including data that are irregularly spaced with non-uniform time intervals; 
can learn higher resolution solutions through learning continuous temporal dynamics from flow field snapshots; and can correctly predict trajectories starting at different initial conditions on which the model was not trained. We also demonstrate that our convergence test method is generally applicable to ML models.
In addition, we derive theoretical error bounds for simple linear ODEs.
Our results show promise in bridging between ML methodologies and scientific computing methodologies, by respecting both the fundamentals of ML and the fundamentals of science.

%% file: _s4_results_comm_physics.tex
\section*{Results and Discussion}
\label{sec:results}

\begin{figure*}[!ht]
	\centering
	\subfloat[Euler-Net] {{ \includegraphics[width=0.2\linewidth]{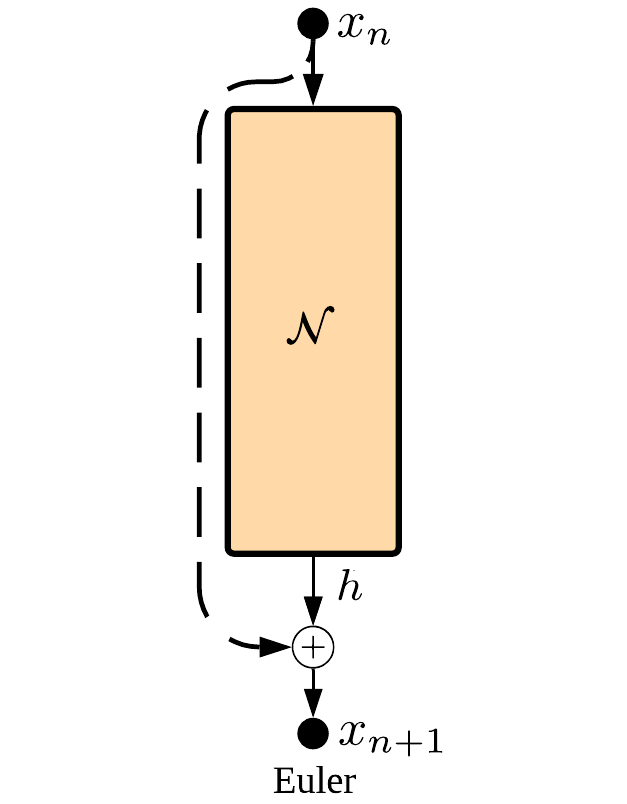} \label{subfig:schematic_euler_net}
	}}
	\subfloat[Euler-Net convergence test] {{ \includegraphics[width=0.33\linewidth]{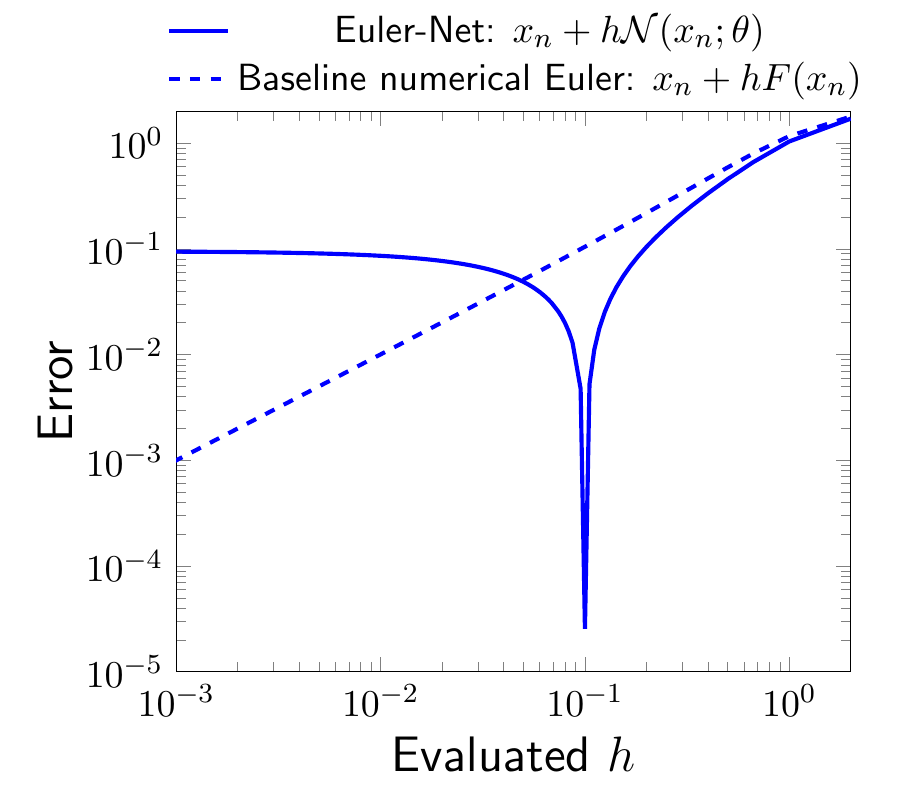} \label{subfig:harm_osc_inference_euler}
	}} 
	\subfloat[Evaluated $h$ is 10\% of the trained $\Delta t$] {{ \includegraphics[width=0.33\linewidth]{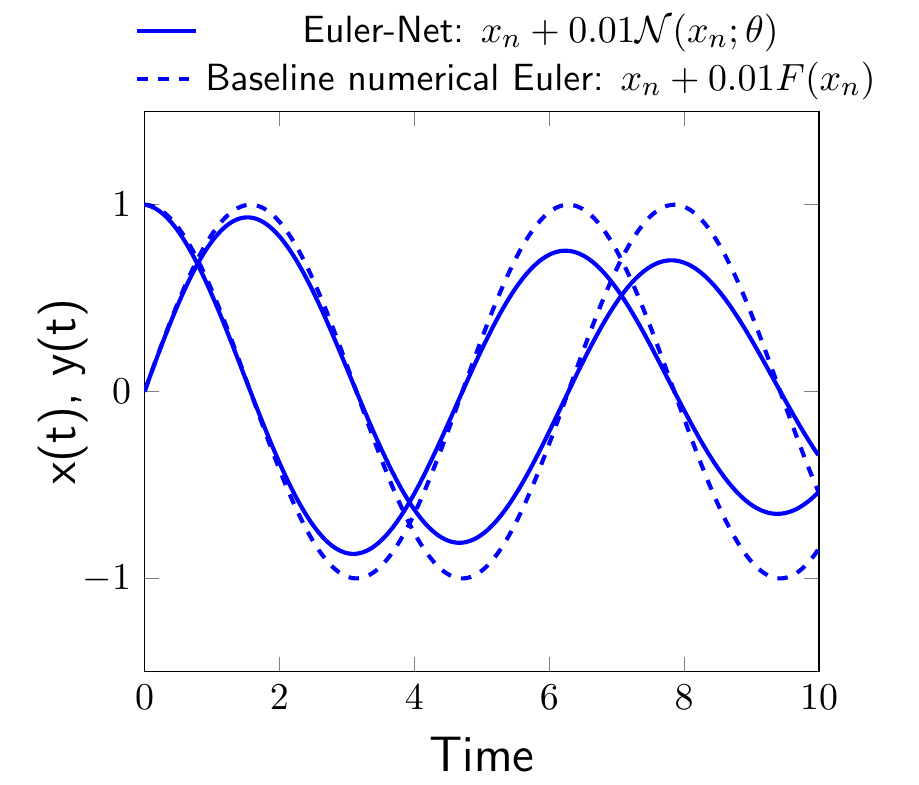}
	\label{subfig:harm_osc_traj_euler}
	}}
	
	\subfloat[RK4-Net] {{ \includegraphics[width=0.2\linewidth]{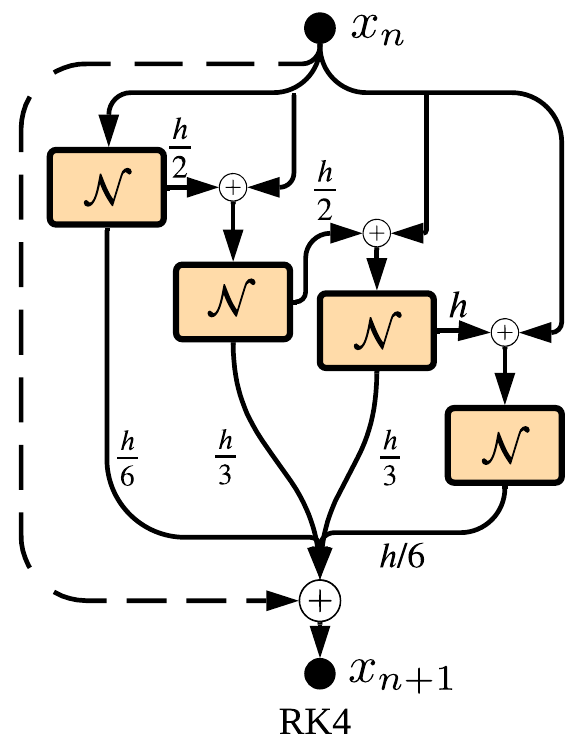}
	\label{subfig:schematic_rk4_net}
	}} 
	\subfloat[RK4-Net convergence test] {{ \includegraphics[width=0.33\linewidth]{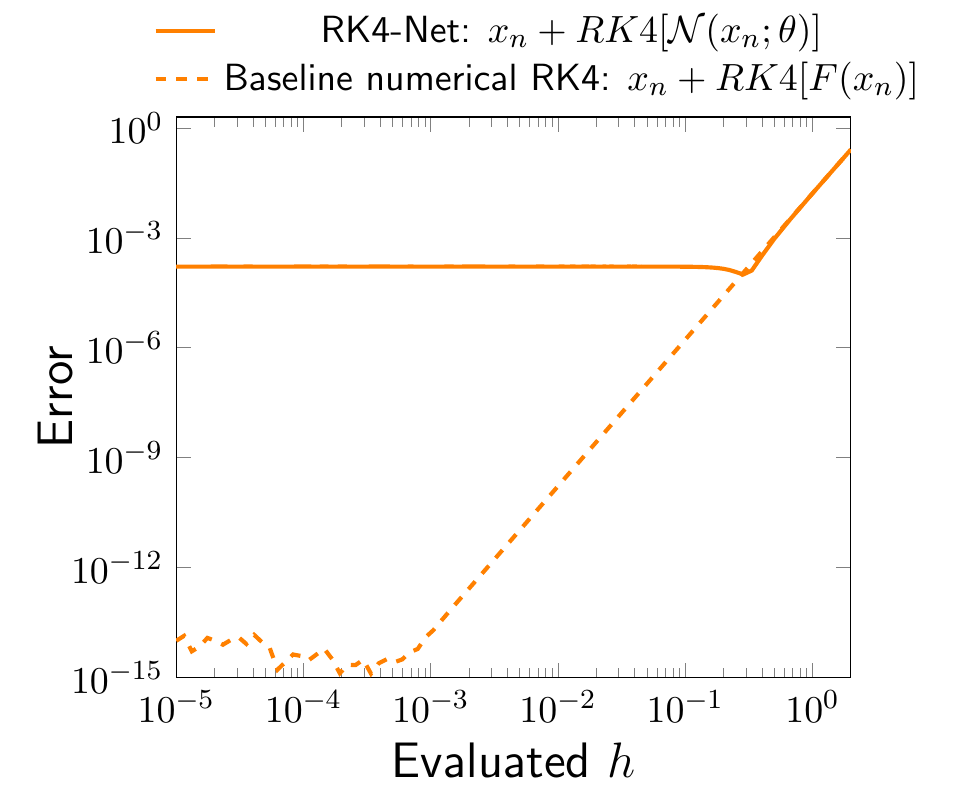}
	\label{subfig:harm_osc_inference_rk4}
	}}
	\subfloat[Evaluated $h$ is 10\% of the trained $\Delta t$] {{ \includegraphics[width=0.33\linewidth]{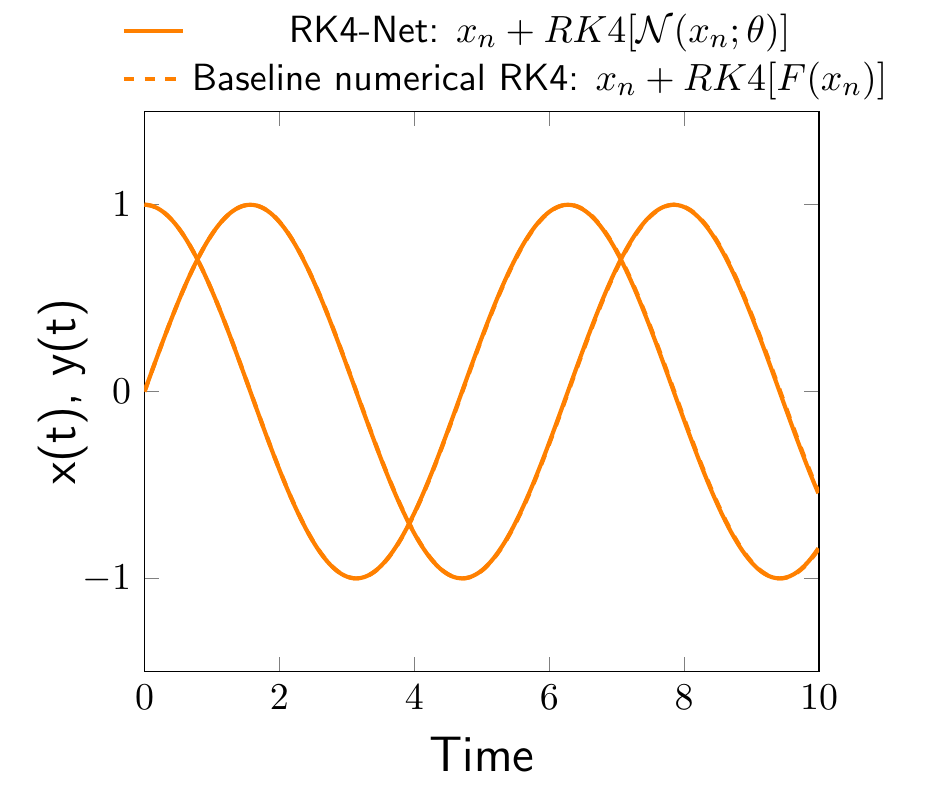}
	\label{subfig:harm_osc_traj_rk4}
	}} \
	\caption{\textbf{Illustration of our convergence test with different ODE-Nets.} 
		\footnotesize 
		\protect\subref{subfig:schematic_euler_net} Schematic of an ODE-Net block, where the next timestep is obtained with the Euler method. 
		In \protect\subref{subfig:harm_osc_inference_euler} and \protect\subref{subfig:harm_osc_traj_euler}, an Euler-Net is trained on discrete data points, spaced apart by some $\Delta t$ (in this case, $\Delta t = 0.1$). 
		\protect\subref{subfig:harm_osc_inference_euler} After training at one specific $\Delta t$, Euler-Net is evaluated at many different step sizes, $h$, both larger and smaller than $\Delta t$. 
		The sharp dip at $h = 0.1$ demonstrates that the model achieves low error when h = $\Delta t$, i.e., it is a good discrete model (DiscreteNet) when the evaluated step size is the same as the $\Delta t$ in the training data.
		However, when the model is evaluated at an $h$ even slightly larger than or smaller than $\Delta t$, the error increases sharply; and thus the Euler-Net model does not pass the convergence test. 
		This is in contrast to numerical integrators, where errors decrease monotonically as $h$ decreases.
		\protect\subref{subfig:harm_osc_traj_euler} We visualize a trajectory where the Euler-Net is evaluated at a $h$ one order of magnitude lower ($h = 0.01$) than the trained $\Delta t$. The resulting trajectory shows that the Euler-Net is unable to follow the baseline numerical Euler solution. 
		\protect\subref{subfig:schematic_rk4_net} Schematic of an ODE-Net block, in which the RK4 numerical integration scheme is used to obtain the next timestep. 
		As before, the RK4-Net is trained on discrete data points. 
		This time, as \protect\subref{subfig:harm_osc_inference_rk4} shows, the error converges monotonically to a fixed value as $h$ decreases; and thus the RK4-Net model passes the convergence test, i.e., it is a good continuous (ContinuousNet) model. 
		The reason the RK4-Net error flattens and converges to a (non-zero but small) fixed value is due to ML-based error sources~\cite{bottou2007}. 
		The RK4 numerical integration scheme also converges to a (non-zero but very small) fixed value---this time, due to numerical-based round-off errors~\cite{chaitin1996lectures}. 
		In \protect\subref{subfig:harm_osc_traj_rk4}, the RK4-Net follows behavior similar to the RK4 numerical integration scheme when evaluated at a low $h$ ($h < \Delta t$). 
	}
	\label{fig:convergence_test_schematic}
\end{figure*}


In this section, we use the convergence test to demonstrate and identify discrete-overfitting of dynamics models.
We start by showing an example of our convergence test on a simple harmonic oscillator system (\nameref{subsec:example_convergence_test}). We then illustrate our convergence test on a variety of different scientific systems, demonstrating that our method can validate whether a trained dynamics model has learned (some) meaningfully continuous dynamics (\nameref{subsec:additional_systems}). 
Next, we show that models that pass this test can
predict fine-scale solutions from coarsely spaced data. 
This includes: 
predicting continuous temporal dynamics from flow fields (\nameref{subsec:gyre_flow}); 
predicting trajectories starting at initial conditions on which the model was not trained (\nameref{subsec:other_initial_conditions}); and
predicting fine-scale solutions from coarse, irregularly spaced data~(\nameref{subsec:irregular_data}). 
We then show that overfitting to the temporal discretization affects ML methods more generally than just with ODE-Nets~(\nameref{sec:sindy}).

Here, we only consider systems which are non-chaotic, non-divergent, and that are not extremely stiff, such that they can be handled by simple explicit Runge-Kutta integrators.
More generally, learning the underlying ``true'' dynamics would require a test that involves a more sophisticated coupling of numerical and ML methodologies.
This is not necessarily needed for many scientific ML tasks, including any of the improvements for predicting fine-scale solutions from coarsely spaced data that we discuss.

\subsection*{Example Convergence Method}
\label{subsec:example_convergence_test}

We demonstrate our ContinuousNet convergence test on a toy example. 
We sample discrete training data points from the linear differential equations describing the harmonic oscillator:
\begin{equation}
\begin{split}
\frac{\mathrm{d}x}{\mathrm{d}t}  = y; \quad 
\frac{\mathrm{d}y}{\mathrm{d}t}  = -x.
\end{split}
\label{eq:eqn_harmonic_oscillator}
\end{equation}
We show the results in \fref{fig:convergence_test_schematic}. Two ODE-Nets are trained on this data. Here, \fref{fig:convergence_test_schematic}\protect\subref{subfig:schematic_euler_net}\protect\subref{subfig:harm_osc_inference_euler}\protect\subref{subfig:harm_osc_traj_euler} use the forward Euler integration scheme, while \fref{fig:convergence_test_schematic}\protect\subref{subfig:schematic_rk4_net}\protect\subref{subfig:harm_osc_inference_rk4}\protect\subref{subfig:harm_osc_traj_rk4} use the RK4 integration scheme.
%
Both the Euler-Net (\fref{fig:convergence_test_schematic}\protect\subref{subfig:schematic_euler_net}) and RK4-Net (\fref{fig:convergence_test_schematic}\protect\subref{subfig:schematic_rk4_net}) use the same linear network architecture $\mathcal{N}(x;\theta)=\theta x$ to approximate the ODE. In theory, a linear model can exactly represent the linear ODE; however, we will demonstrate that this does not happen.
For the example in \fref{fig:convergence_test_schematic}, the training data was generated from the analytical solution, spaced apart by the training timestep $\Delta t = 0.1$.
To measure the performance as a continuous model at inference time, we integrate the model using a range of inference timesteps $h$.
%
Figures~\ref{fig:convergence_test_schematic}\protect\subref{subfig:harm_osc_inference_euler} and \ref{fig:convergence_test_schematic}\protect\subref{subfig:harm_osc_inference_rk4} plot the results of the convergence test with the Euler-Net and the RK4-Net.

For the Euler-Net, the error when $h = \Delta t$ (the step size is equal to the temporal spacing in the training data) is very low, but it increases when $h$ decreases. 
This is in contrast to the classical Euler numerical integration scheme, where the error decreases as $h$ decreases. 
Thus, these results for Euler-Net do \emph{not} pass the convergence test. 
In contrast, for the RK4-Net, the error decreases as $h \rightarrow 0$, and eventually it approaches and levels off at a fixed value. 
Notably, the error does not increase dramatically as it does with the Euler-Net. 
In this case, the RK4-Net has learned the right inductive biases to approximate an underlying continuous dynamics for the system.

We illustrate this further by showing an example trajectory at a specific evaluated $h$. 
In this case, both trained ODE-Nets are evaluated at $h = 0.01 $ (a $10\times$ increase in resolution in comparison to the training data) up to a final timestep. 
In \fref{fig:convergence_test_schematic}\protect\subref{subfig:harm_osc_traj_euler}, the Euler-Net falls off of the true numerical Euler solution. 
It has clearly not learned the underlying continuous dynamics. 
In contrast, in \fref{fig:convergence_test_schematic}\protect\subref{subfig:harm_osc_traj_rk4}, the RK4-Net shows good correspondence with the true numerical RK4 solution.

\subsection*{Four Prototypical Dynamical Systems}
\label{subsec:additional_systems}
We consider canonical scientific dynamical systems: the non-linear pendulum, the Lotka-Volterra equations, the Cartesian pendulum, and the double gyre fluid flow.
The first two systems are non-linear dynamical systems; the Cartesian pendulum is a stiff dynamical system (which is difficult to solve with numerical methods without taking very small timesteps); and the double gyre fluid flow consists of vorticity fields describing a stream function. We provide more details about these dynamical systems in~\nameref{sec:ds_systems}.
For each system, we sample data points from either the analytical solution or the numerical solution. The temporal spacing between the discrete data points is denoted as $\Delta t$, while the step size used to evaluate a trained ODE-Net is denoted as~$h$.

\paragraph{Training setup.}
We train an ODE-Net with a numerical integration scheme (Euler or RK4) for each system. 
We use simple feed-forward networks with tanh activation functions. 
See~\nameref{sxn:app_architecture} for details on the architecture used.
In every example, the exact same network architecture is used for both the Euler-Net and RK4-Net, respectively. 
We also include additional results for ODE-Nets trained on training data spaced apart by different $\Delta t$ as well as an ODE-Net trained with the Midpoint numerical integration scheme, in~\nameref{sec:appendix_additional_examples_convergence}. For the double gyre fluid flow, we use a dynamic autoencoder architecture~\cite{erichson2019physics,azencot2020forecasting} to embed the high-dimensional input of flow field snapshots in some latent space. Specifically, we replace the linear discrete map in the architecture proposed by~\cite{erichson2019physics} with a linear ODE-block. This means that the model learns to predict the next timestep by integrating forward in latent space (using an Euler or RK4 numerical integration scheme) with step size $h = \Delta t$. Finally, the decoder translates the latent space vectors back to the flow field.

\begin{figure}[!t]
	\centering
	\subfloat[Non-linear pendulum] {{ \includegraphics[width=0.33\linewidth]{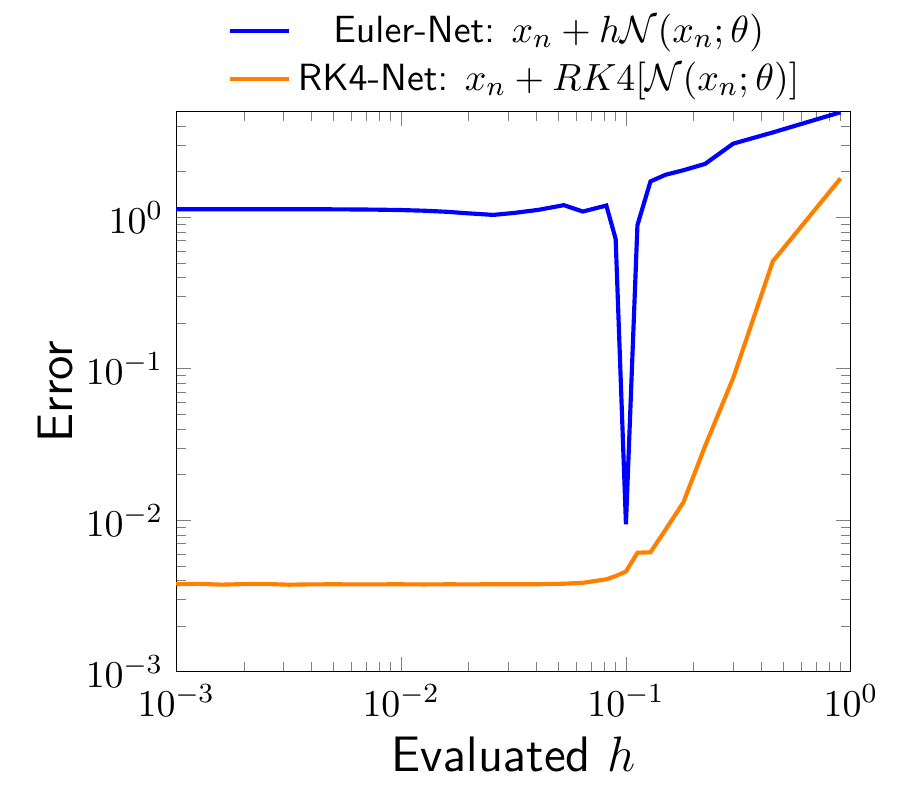}
			\label{subfig:nonlinear_pendulum_convergence}
	}}
	\subfloat[Lotka-Volterra system] {{ \includegraphics[width=0.33\linewidth]{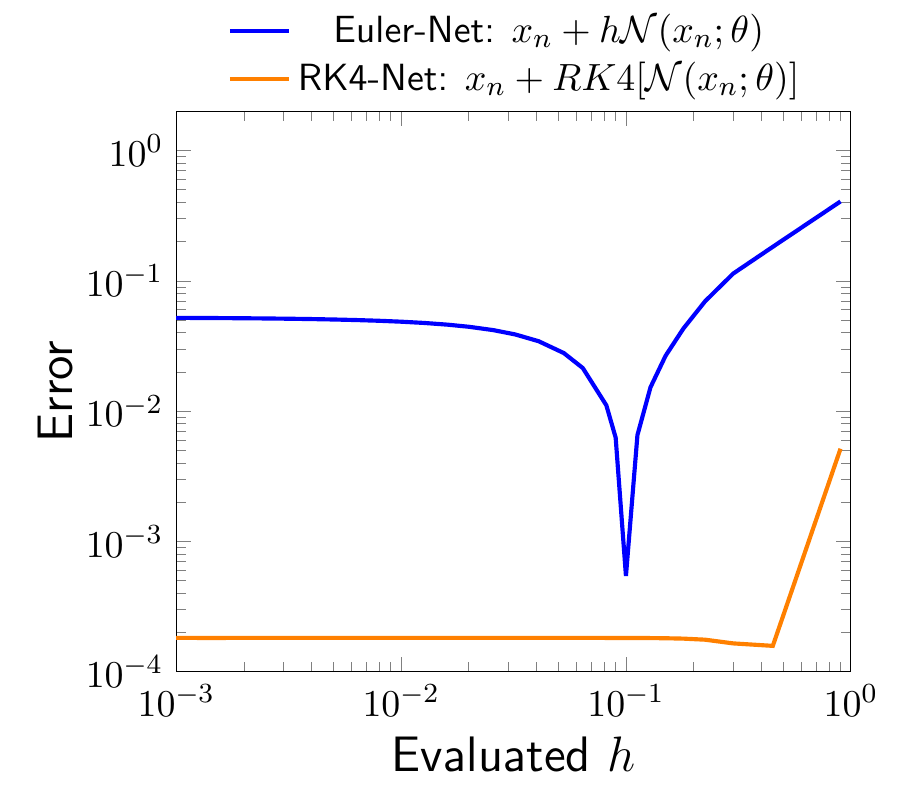}
			\label{subfig:lotkavolterra_convergence}
	}} 
	
	\subfloat[Cartesian pendulum] {{ \includegraphics[width=0.33\linewidth]{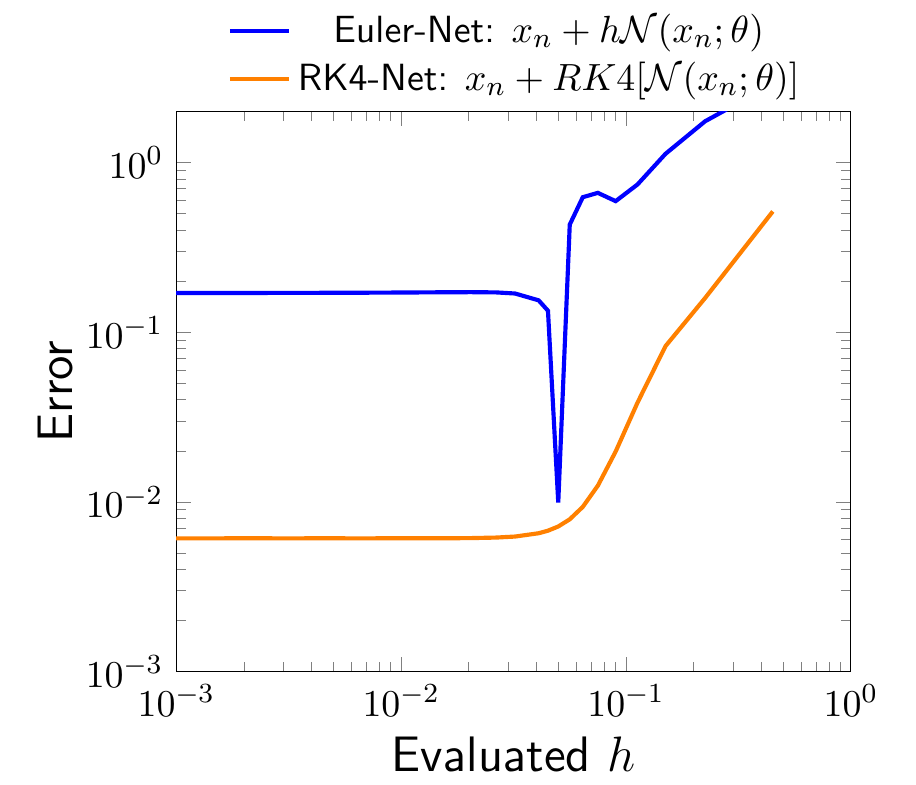}
			\label{subfig:cartesian_convergence}
	}} 
	\subfloat[Double gyre fluid flow] {{ \includegraphics[width=0.33\linewidth]{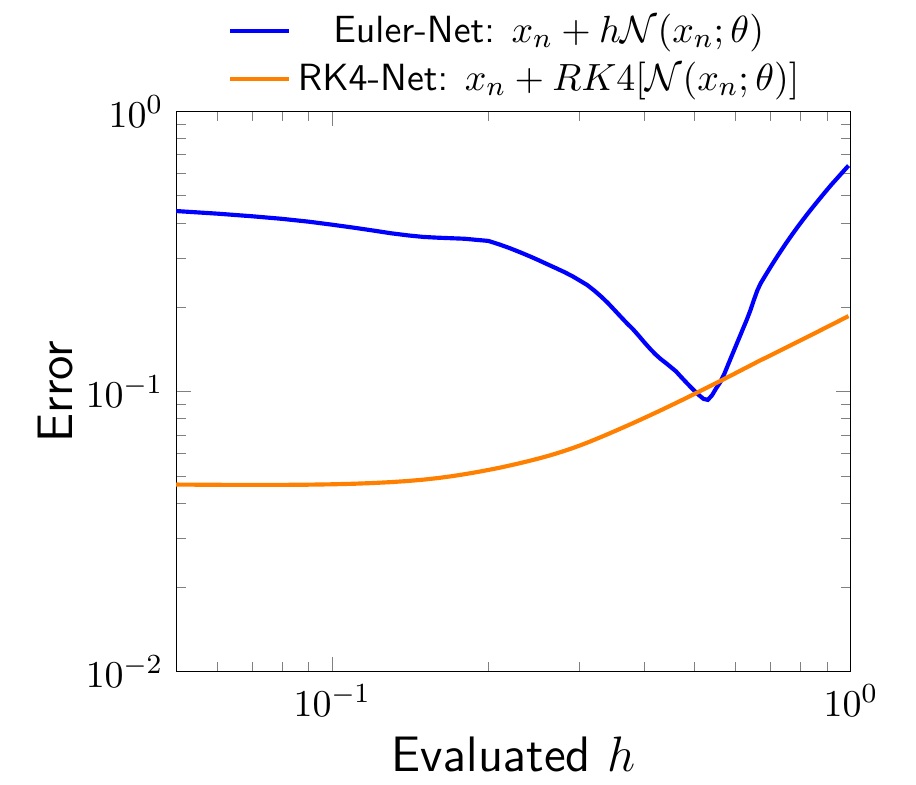}
			\label{subfig:fluidflow_convergence}
	}} 
	\caption{\textbf{Illustration of our convergence test on prototypical dynamical systems.}  We demonstrate the convergence test on multiple systems: (a) Non-linear pendulum (b) Lotka-Volterra system (c) Cartesian pendulum (d) Double gyre fluid flow. In each case, for the Euler-Net, the error does not monotonically decrease: error is low when $h = \Delta t$, but high at all other $h$ evaluations. Thus, the Euler-Net does not pass the convergence test. In contrast, the RK4-Net error does monotonically decrease and converges to a fixed value, when the evaluated $h \rightarrow 0$. The RK4-Net does pass the convergence test.
	}
	\label{fig:euler_rk4_convergence_trajectory_multiple}
\end{figure}

\paragraph{Results.}
The results of our method are shown in~\fref{fig:euler_rk4_convergence_trajectory_multiple}. 
In each case, the Euler-Net has low error when $h = \Delta t$ (i.e., evaluated at the same time spacing as the training data), but it has high error when evaluated at all other $h$, in particular smaller values of $h$. 
Thus, it does not pass the convergence test, and it has not learned a meaningfully continuous dynamics.
It is a good discrete model, appropriate for data drawn from the same temporal discretization, but it has overfit to the temporal discretization. 
In contrast, the error during inference time of the RK4-Net steadily decreases when it is evaluated at lower $h$, eventually converging to a fixed basal level determined by the model and the noise properties of the data. It has passed the convergence test, and it can be said to have learned a meaningfully-continuous model. We include additional convergence test results in Supplementary Figures 9-12.

\subsection*{Interpolation: Predicting Fine-scale Solutions from Coarse Training Data}
\label{subsec:gyre_flow}

\begin{figure}[t] 
\centering 
\vspace{1ex}
{{ \begin{overpic}[width=0.70\columnwidth]{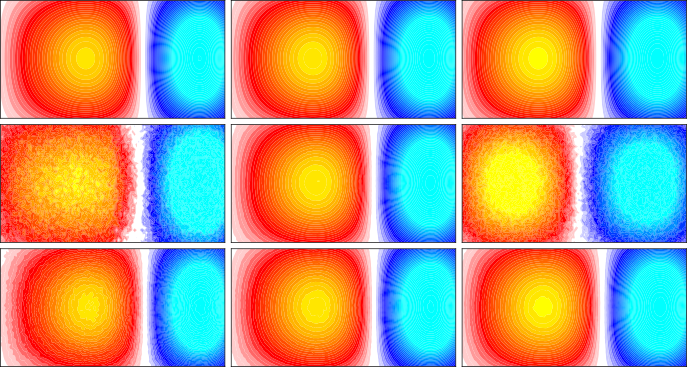}
    \put(12,54.5){\footnotesize $h=0.25$}
    \put(43,54.5){\footnotesize$h = \Delta t=0.5$}
    \put(78,54.5){\footnotesize $h =0.8$}
    
    \put(-3,42){\rotatebox{90}{\footnotesize Truth}}
    \put(-3,22){\rotatebox{90}{\footnotesize Euler-Net}}
    \put(-3,4){\rotatebox{90}{\footnotesize RK4-Net}}
   \end{overpic}
}}

\caption{\textbf{Double gyre fluid flow: Reconstructing fine-scale flow fields from coarse training data.} The training data for this problem consists of vorticity field snapshots of the dynamical system taken at $\Delta t =0.5$. In the images above, the red region is rotating in one direction, and the blue region is rotating in the opposite direction. After training an Euler-Net and an RK4-Net,
both models are evaluated at different $h$ (both when $h > \Delta t$ and $h < \Delta t$) at a final timestep. We show the evaluation results for this final timestep, $T$, at $h = 0.25$, $h = 0.5 = \Delta t$, and $h = 0.8$. 
The Euler-Net (which fails the convergence test) approximates a solution close to the true solution at a timestep in the training data but does poorly at the other timesteps; it does not capture the fluid flow behavior and gives a grainer solution. The RK4-Net (which passes the convergence test) is able to output solutions that have close correspondence to the true solution; it successfully interpolates the fine-scale flow fields that are in-between the training data snapshots, resulting in a much higher frame rate~solution.}

\label{fig:gyre_fluid_flow}
\end{figure}

Observational, discrete training data are limited in that they are measured at specific timesteps. 
To obtain a solution for the system in-between these timesteps, one must retake the data measurements again at finer timesteps. 
However, selecting a model that has learned meaningfully continuous dynamics should guarantee accurate evaluation at smaller timesteps, despite only training on coarse and/or irregularly spaced temporal data (i.e., measurements taken with large timesteps). 
By learning continuous dynamics, the trained ODE-Net model can be evaluated at any point in temporal space, and still yield a low error solution.
In this case, one would not need to recollect training data with smaller $\Delta t$ between data points; the learned ODE-Net can be used instead.
Here, we demonstrate that fine-scale evaluation is possible by learning continuous temporal dynamics from flow fields for the double gyre flow example.

\paragraph{Results.} 
We consider two models: the Euler-Net which did not pass the convergence test, and the RK4-Net which did pass the convergence test (see~\fref{fig:euler_rk4_convergence_trajectory_multiple}). 
In~\fref{fig:gyre_fluid_flow}, we show the flow field snapshots that result from both models being evaluated at different timesteps. 
The Euler-Net is only able to approximate the true solution at the training data timestep (in this case, $h = \Delta t = 0.5$). 
It cannot match the true solution at the other timesteps, and it gives a poor approximation that does not capture the flow behavior. 
In contrast, the RK4-Net has good correspondence to the true solution even when it is evaluated at timesteps that were not in the training data. 
Thus, our convergence test method has allowed us to choose a model that can recover fine-scale solutions of the system, while only having access to coarse-scale measurements during training.

\subsection*{Extrapolation: Predicting Trajectories for New Initial Conditions}
\label{subsec:other_initial_conditions}

\begin{figure}[!t]
	\centering
	\subfloat[Phase portrait (true solution)] {{ \includegraphics[width=0.33\linewidth]{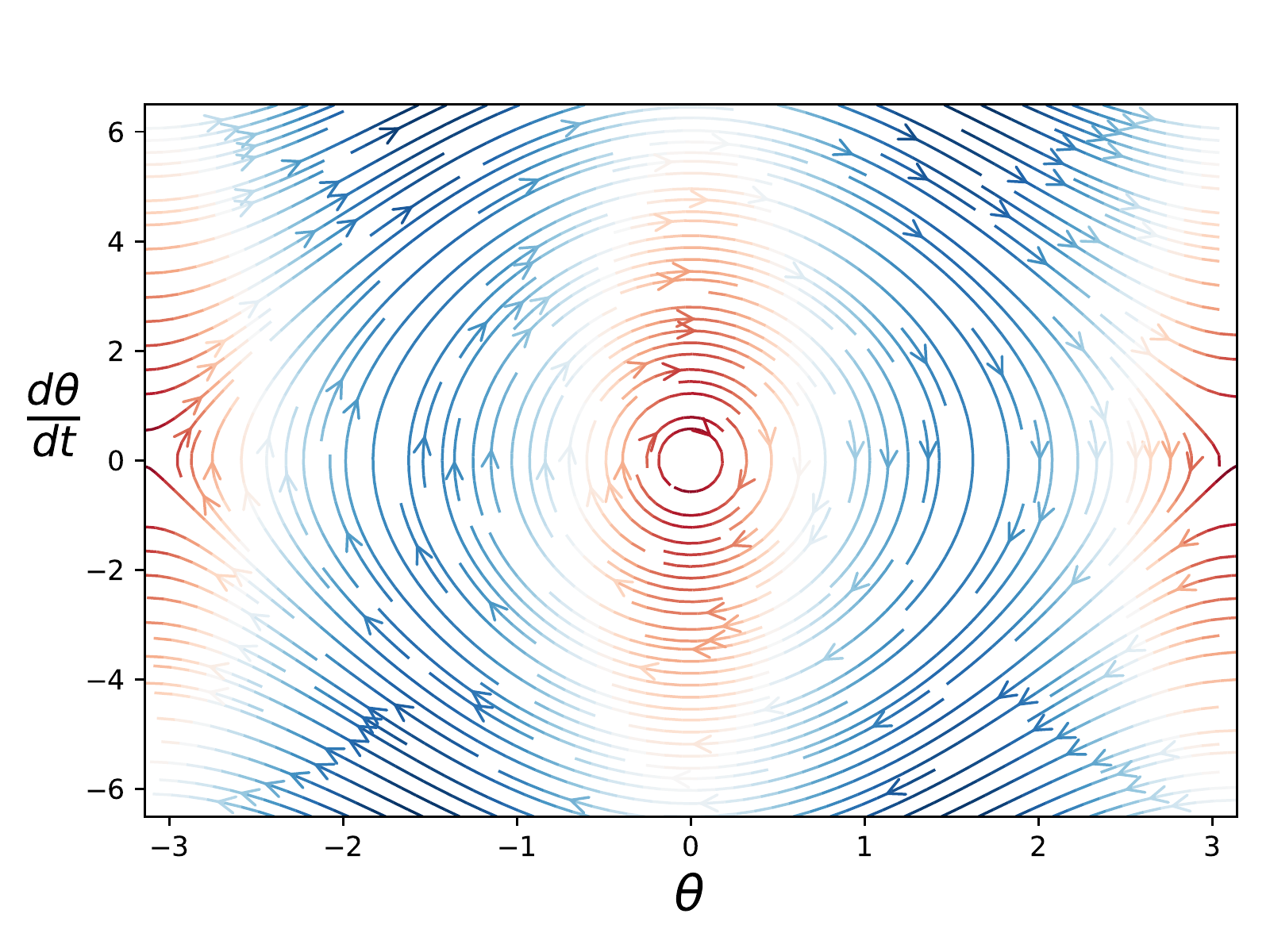}
			\label{subfig:extrapolate_true}
	}} 
	\subfloat[Training data] {{ \includegraphics[width=0.33\linewidth]{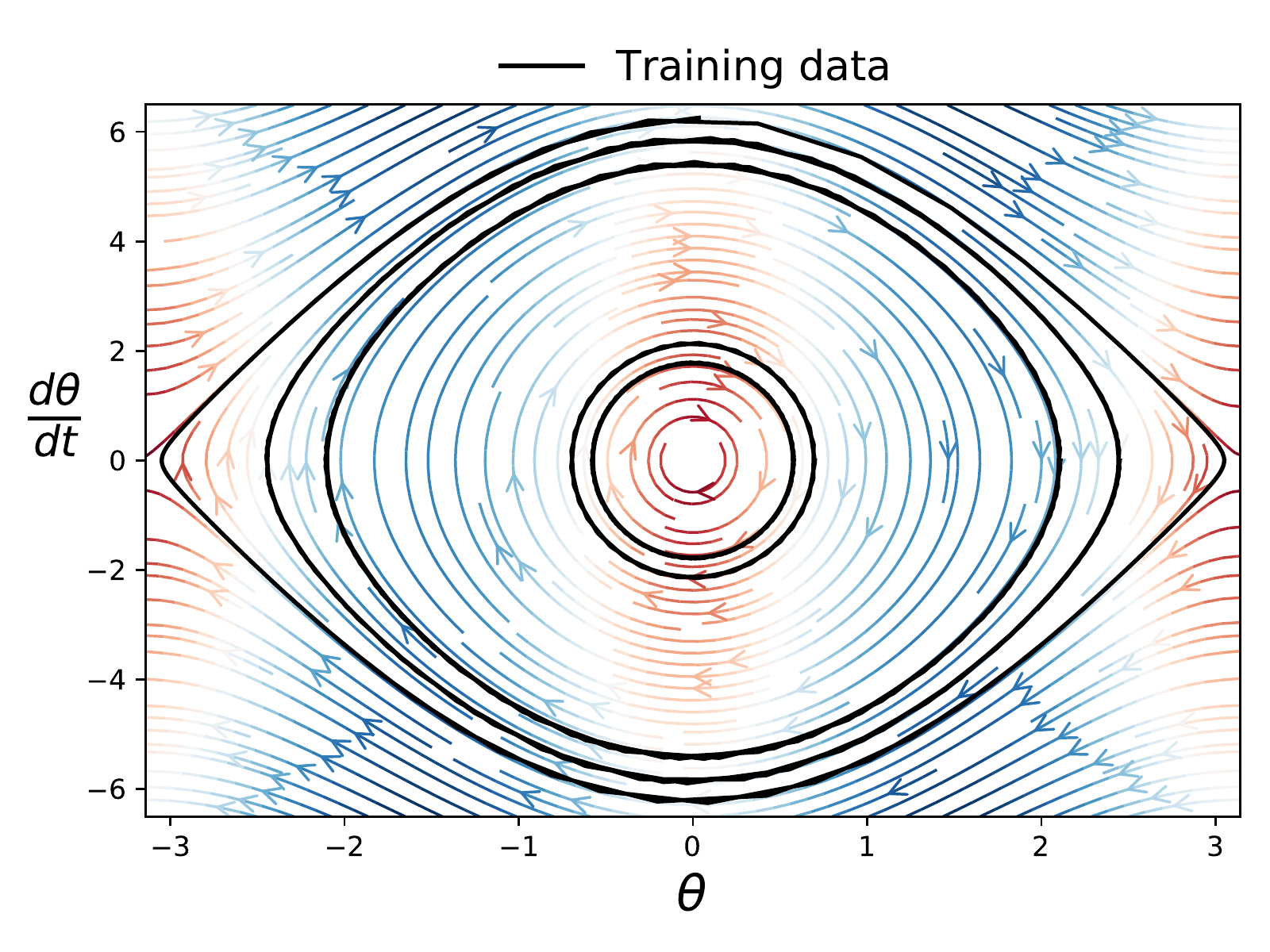}
			\label{subfig:extrapolate_training}
	}}
	\vspace{-1ex}
	\subfloat[Incorrect model (Euler-Net) ] {{ \includegraphics[width=0.33\linewidth]{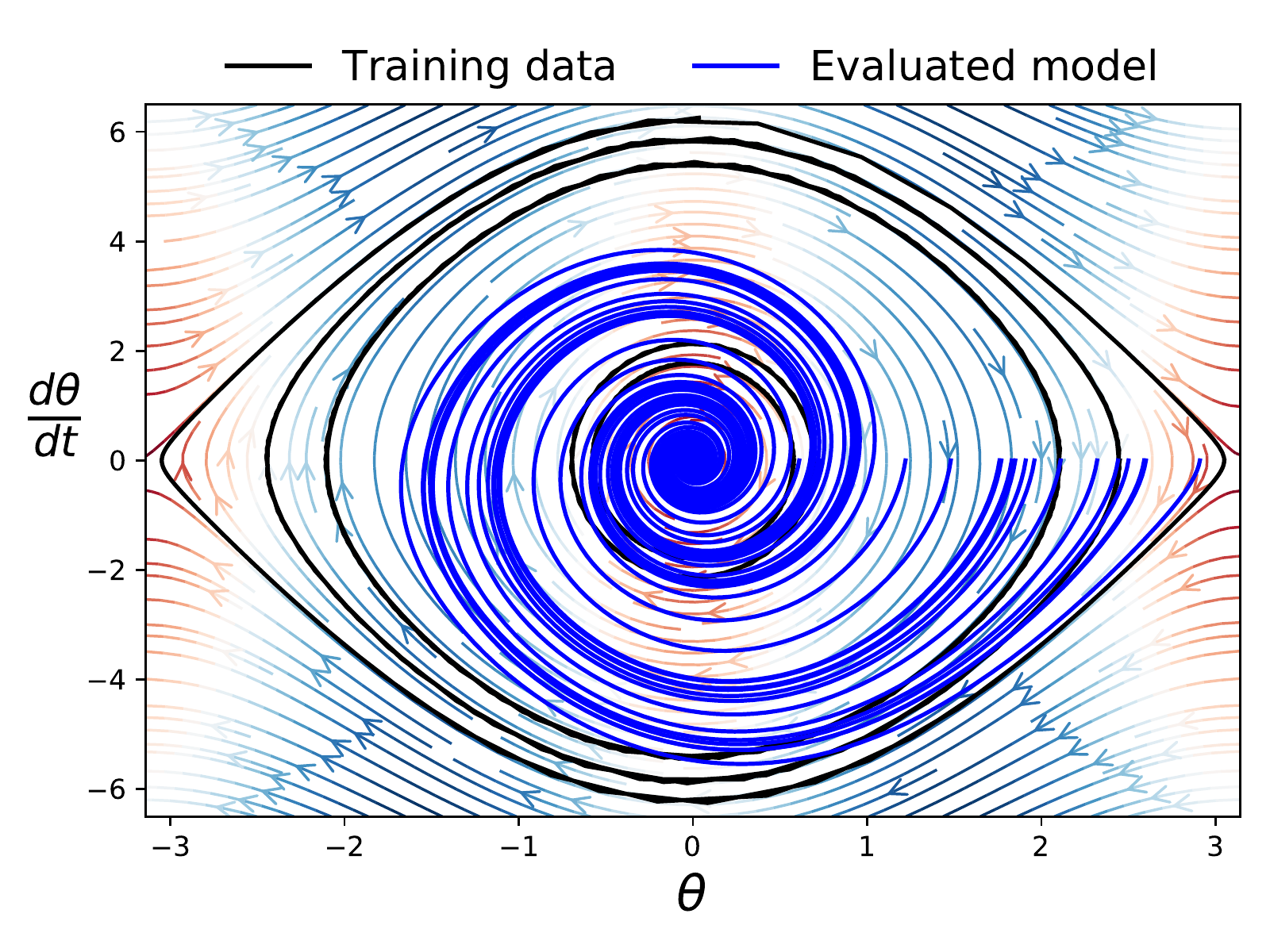}
			\label{subfig:extrapolate_euler}
	}} 
	\subfloat[Correct model (RK4-Net)] {{ \includegraphics[width=0.33\linewidth]{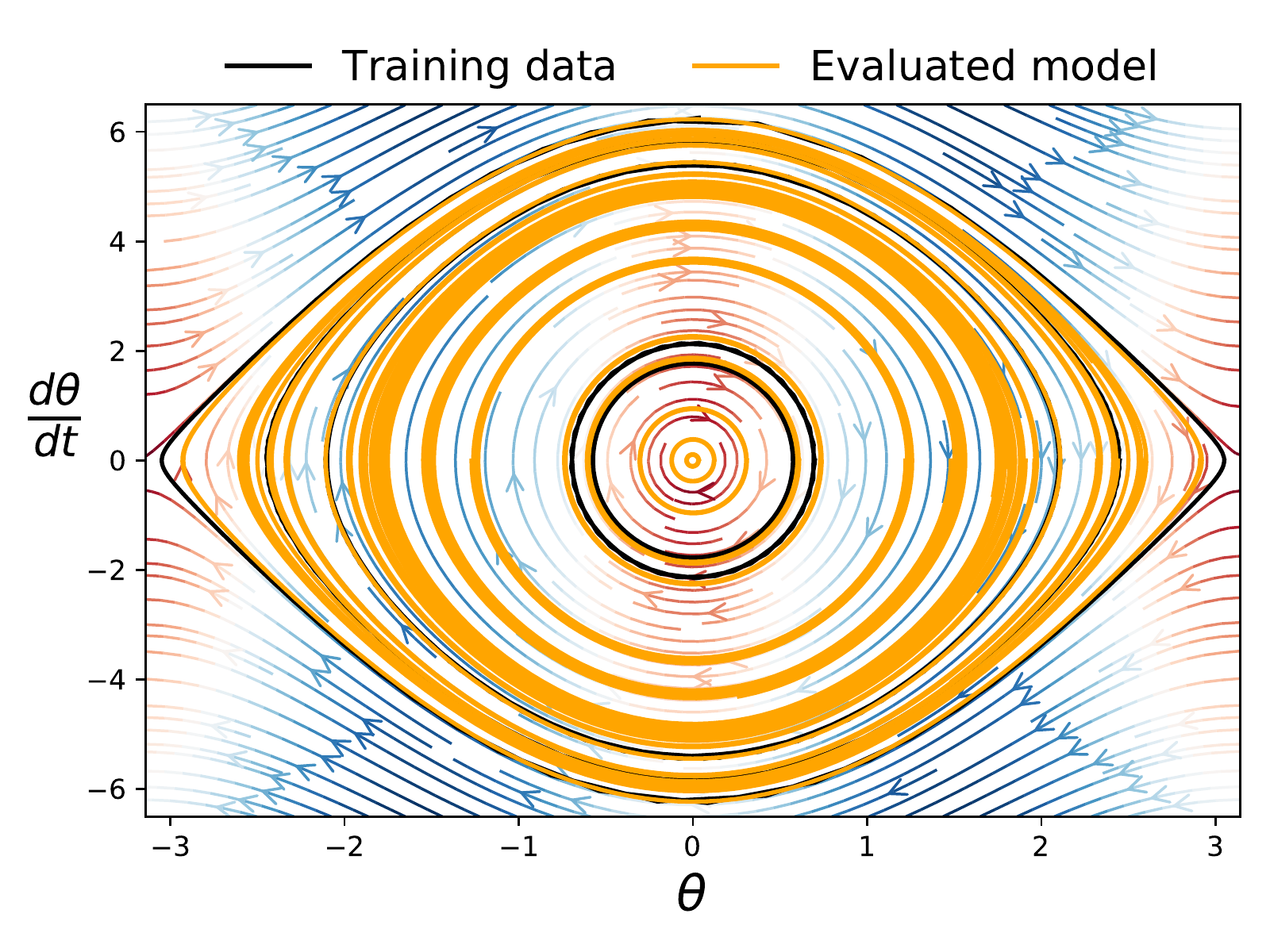}
			\label{subfig:extrapolate_rk4}
	}}
	\caption{\textbf{Non-linear pendulum: Extrapolation to predict initial condition trajectories on which the model was not trained.} 
		ODE-Net models are trained on randomly chosen initial conditions (different $\theta$ values), which have temporal points spaced apart by $\Delta t = 0.1$ (see \protect\subref{subfig:extrapolate_training}). Each  model is then evaluated on a different set of initial conditions (not in the training data) at much smaller step sizes ($h = 0.001$, $100\times$ higher resolution). After evaluation, the Euler-Net quickly falls off of the phase plot lines corresponding to the true solution for the different test trajectories. The RK4-Net, which has learned a continuous dynamics, is able to extrapolate to different trajectories (starting at different initial conditions), with good correspondence to the phase plot lines of the true solution.
	}
	\label{fig:pendulum_general}
\end{figure}

For a given system, temporal trajectories start at some initial condition. 
Measurements are taken for one trajectory at one initial condition, and then must be taken separately for other trajectories with different initial conditions. 
Selecting a model that has learned a meaningfully continuous dynamics circumvents this: after training a model on data points sampled from one (or more) trajectories, the model should be able to extrapolate and predict accurate solutions for new initial~conditions.

\paragraph{Training setup.} 
We look at the non-linear pendulum (described in~\eref{eq:non_linear_pendulum} in~\nameref{sec:ds_systems}). 
Here, $\theta$ is the initial condition representing the position of the pendulum in time. The phase portrait of this system (representing the true solution trajectories), showing $\frac{\mathrm{d} \theta}{\mathrm{d}t}$ against $\theta$, is shown in~\fref{fig:pendulum_general}\protect\subref{subfig:extrapolate_true}. 
An Euler-Net and an RK4-Net are trained on trajectories, spaced apart by $\Delta t = 0.1$, starting at certain initial conditions (shown by the black lines in~\fref{fig:pendulum_general}\protect\subref{subfig:extrapolate_training}). We then pick a test set of a number of different initial conditions that were not in the training data. The Euler-Net and RK4-Net start at these test initial conditions and are both evaluated at a finer $h  = 0.001$, representing a $100\times$ increase in resolution. Note that we saw in~\fref{fig:euler_rk4_convergence_trajectory_multiple} that the Euler-Net did not pass the convergence test (i.e., it had high error when evaluated at $h \ll \Delta t$), while the RK4-Net did pass the test.

\paragraph{Results.} 
The results of predicting trajectories starting at different test initial conditions are shown in~\fref{fig:pendulum_general}. 
The Euler-Net is unable to predict these trajectories and quickly falls off of the phase plot lines corresponding to the true solution. 
In contrast, the RK4-Net is able to predict the trajectories, starting at different test initial conditions with good correspondence to the true solution. 
Thus, we see that it is critical to find a model that passes the convergence test and is able to learn a continuous dynamics to succeed at this extrapolation task.

\begin{figure}[!t]
	\centering 
	\subfloat[$\Delta t$ of training data distribution] {{ \includegraphics[width=0.305\linewidth]{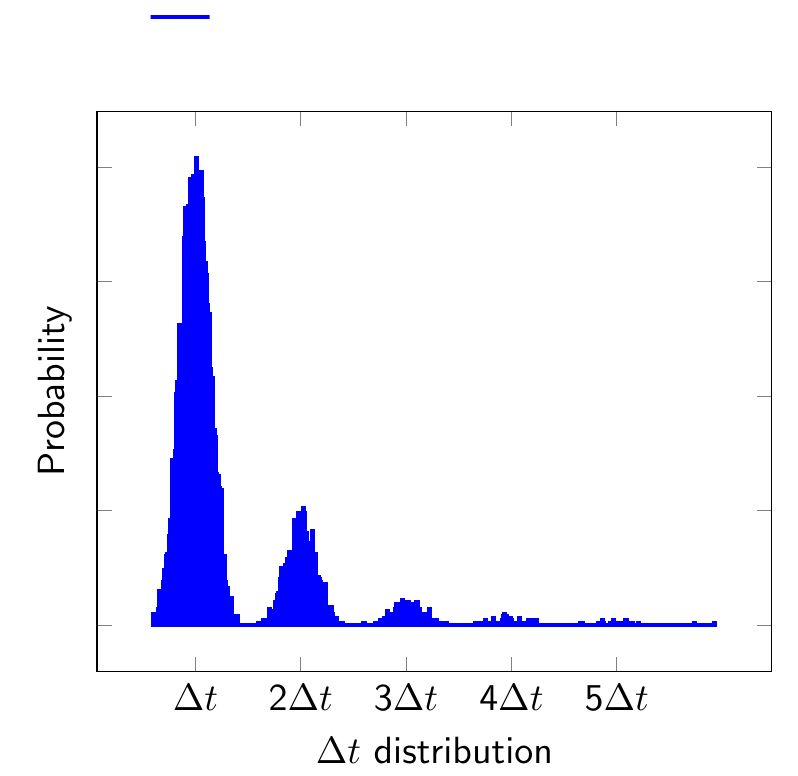}
			\label{subfig:training_data_distribution}
	}} 
	\subfloat[Fine-scale evaluation] {{ \includegraphics[width=0.33\linewidth]{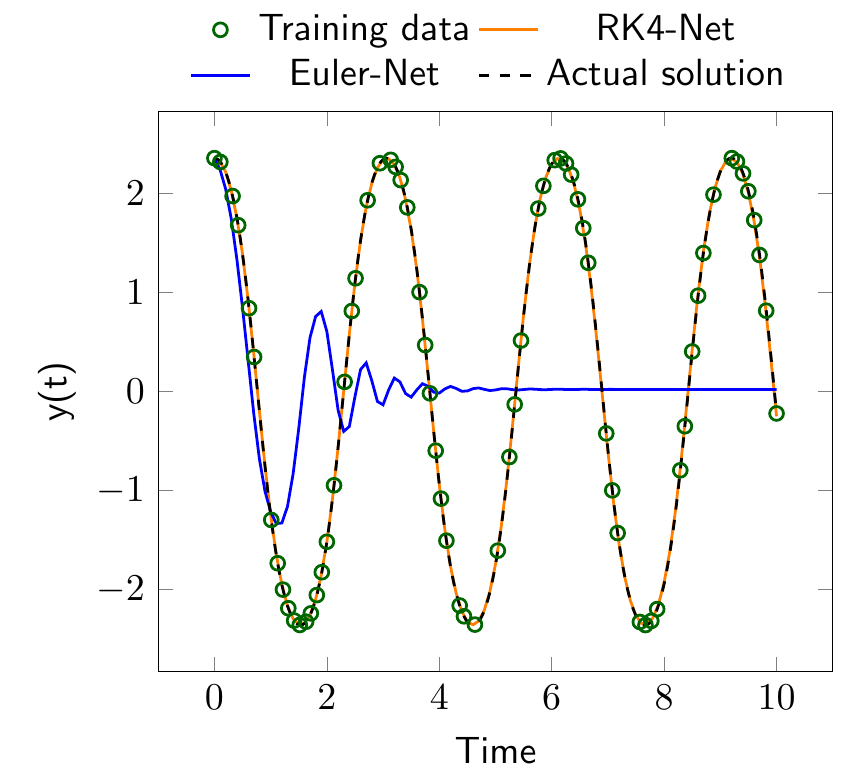}
			\label{subfig:finescale_eval}
	}}
	\vspace{-1ex}
	\subfloat[Zoomed in fine-scale evaluation] {{ \includegraphics[width=0.33\linewidth]{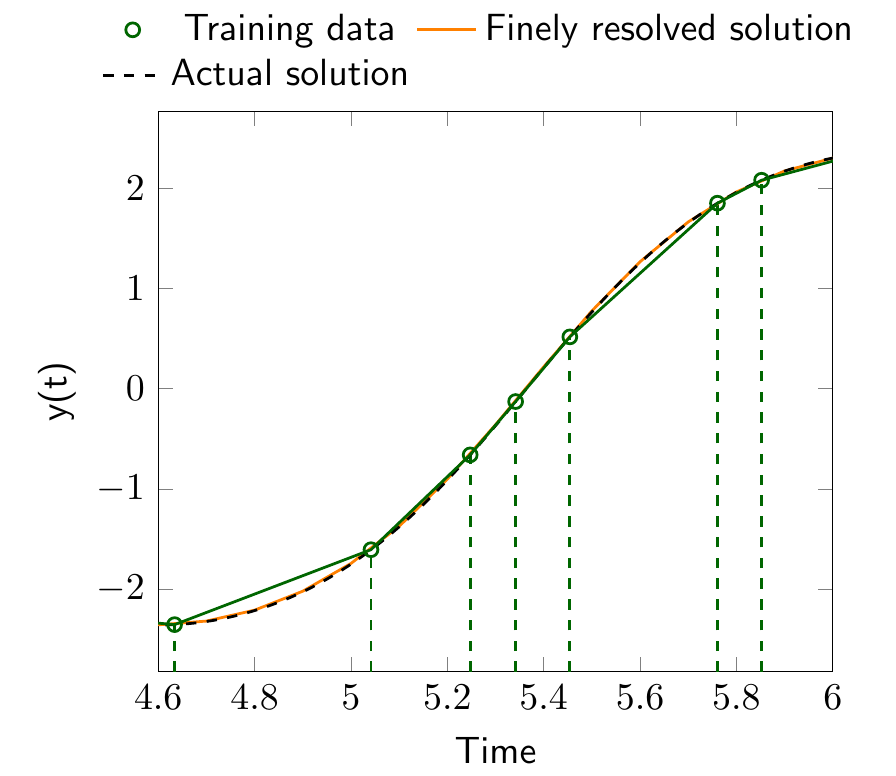}
			\label{subfig:zoomed_finescale_eval}
	}} 
	\subfloat[Convergence test] {{ \includegraphics[width=0.33\linewidth]{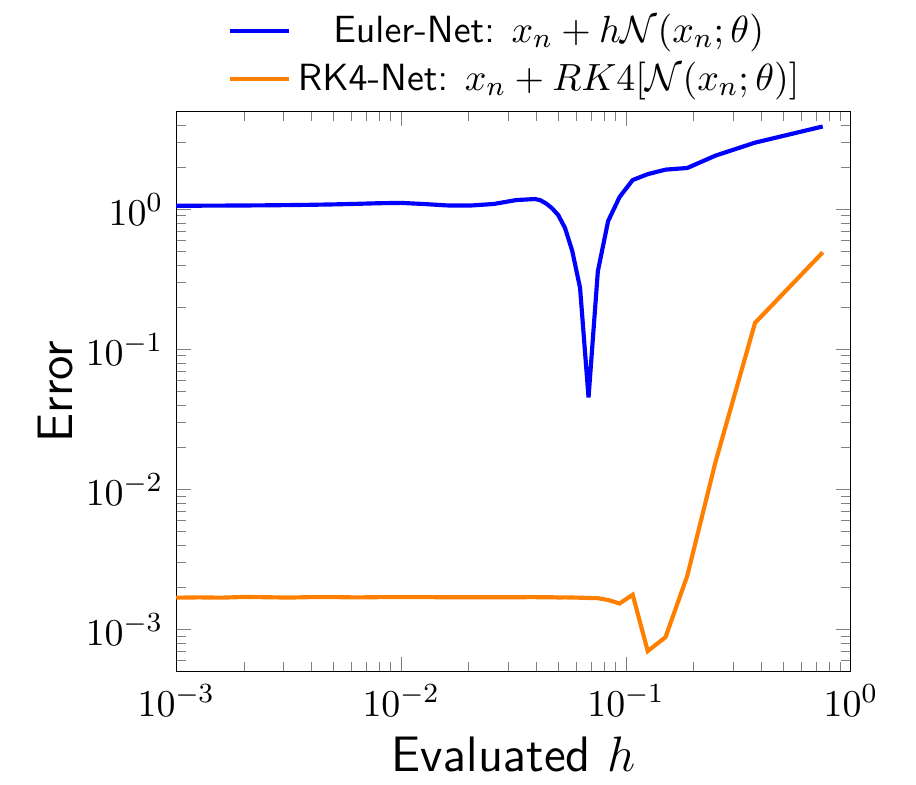}
			\label{subfig:irregular_convergence_test}
	}}
	\caption{\textbf{Learning continuous dynamics from irregularly spaced discrete points.} 
		\protect\subref{subfig:training_data_distribution} 
		Training data distribution for a scientific problem where temporal data measurements are taken with some amount of imprecision (e.g., the measurement of interest is not always taken at the exact same $\Delta t$) and/or measurements are skipped (e.g., we only have a measurement at $\Delta t$ and $3 \Delta t$ because the measurement at 2$\Delta t$ failed or was lost). 
		\protect\subref{subfig:finescale_eval} Different ODE-Nets are trained on the irregularly spaced data (indicated by green dots), as denoted by the green circles. Then, the trained models are evaluated at a very low $h$ (where $h \ll \Delta t$). The RK4-Net (orange) is able to learn a continuous dynamics and follow the continuous solution over time (indicated by the dashed line), while the Euler-Net (blue) does not. 
		\protect\subref{subfig:zoomed_finescale_eval} The RK4-Net (orange) is able to reconstruct the fine-scale, high resolution solution (indicated by dashed line), with good correspondence to the continuous solution, from the coarse, irregularly spaced training data (indicated by green dots).
		\protect\subref{subfig:irregular_convergence_test} The RK4-Net (orange) passes the convergence test, but the Euler-Net (blue) does not.}
	\label{fig:irregular_example_finescale_evaluation}
\end{figure}

\subsection*{Irregularly Sampled Training Data}
\label{subsec:irregular_data}

It is typically the case that scientific data collection includes measurements that are taken with some amount of imprecision. 
For example, the measurement of interest is not always taken at the exact same $\Delta t$ every time, due to issues such as jitter in the measurement device. 
Measurements may also be skipped: for example, a measurement is only available at $t=\Delta t$ and $t=3\Delta t$ because the measurement at $t=2\Delta t$ was lost or skipped. 
Thus, reconstructing the correct trajectory when the measurements are non-uniformly spaced is important in numerous science and engineering problems. 
Here, we look at an example of using the convergence test to correctly select a meaningfully continuous model in the case of the non-linear pendulum with irregularly spaced temporal data with non-uniform temporal~intervals.

\paragraph{Training setup.} 
An example distribution of irregularly sampled training data is shown in~\fref{fig:irregular_example_finescale_evaluation}\protect\subref{subfig:training_data_distribution}. The baseline $\Delta t$ is 0.05, subject to jitter and frameskipping errors. An Euler-Net and an RK4-Net are both trained on these temporal data points, where the values of $\Delta t$ are input into the integration schemes. (That is, at every given measurement, the timestep jiter was also recorded to use in training.) Both ODE-Nets are then evaluated at a very low $h$ (approximately $100\times$ lower than the general distribution of the training data points) to generate a time series plot. Note, that we run the convergence test on both ODE-Nets.

\paragraph{Results.} 
The Euler-Net quickly falls off of the continuous solution (\fref{fig:irregular_example_finescale_evaluation}\protect\subref{subfig:finescale_eval}). 
Conversely, the RK4-Net follows the continuous solution with good accuracy, including at timesteps not in the training data (\fref{fig:irregular_example_finescale_evaluation}\protect\subref{subfig:zoomed_finescale_eval}). 
Thus, it is clear that the RK4-Net has learned a meaningfully continuous dynamics while the Euler-Net has not. 
This is confirmed by RK4-Net passing the convergence test, but Euler-Net not passing it (\fref{fig:irregular_example_finescale_evaluation}\protect\subref{subfig:irregular_convergence_test}). The dip for Euler-Net appears at the average $\Delta t$ in the training data, which is slightly larger than 0.05 due to the measurement noise.

\subsection*{Sparse Identification of Nonlinear Dynamical Systems}
\label{sec:sindy}
Here, we demonstrate that ovefitting to the temporal discretization affects ML methods (in the context of dynamical systems) more generally. To illustrate this, we consider the SINDy learning approach, which is a class of methods for system identification~\cite{brunton2016discovering}.

The SINDy method uses the following model structure to represent the dynamics, 
    \begin{equation}
        \frac{\mathrm{d}x}{\mathrm{d}t} = \Xi \phi(x) ,
    \end{equation}
where $\Xi$ is a matrix of learnable parameters and $\phi(x)$ is a set of non-linear basis functions which correspond to potential terms in the underlying system.
A linear optimizer is used to fit the parameters $\Xi$ to the data, with a sparsity constraint.
The sparsity constraint identifies the subset of relevant basis elements in $\phi(x)$ to reveal an interpretable dynamics model.

In most real applications, the time derivatives, $\mathrm{d}x_n/\mathrm{d}t$, cannot be measured directly and instead need to be approximated from the observations $x_n$.
The common approach in SINDy is to use finite differences to approximate $\mathrm{d}x_n/\mathrm{d}t$ from the data~\cite{desilva2020, Kaptanoglu2022}. (This treatment of time derivatives, where SINDy differentiates the data, is in contrast to the ODE-Net method, which integrates the model.) The finite difference operator $FD(x_{n...})$ is applied over the dataset as a pre-processing step. This yields the following set of $N$ discrete equations, which is optimized for $\Xi$ over all observations~$n$:
    \begin{equation}
        FD(x_{n...}) = \Xi \phi(x_n) ,
    \end{equation}
    where $FD(x_{n...})$ is a finite difference approximation using the region of points around time index $n$. Using the series of $N$ equations, $\Xi$ is learned using specialized algorithms designed to seek sparsity, such as LASSO regularization or sequential threshold least squares~\cite{brunton2016discovering}.
    (This is again in contrast with the ODE-Net method, which uses gradient descent nonlinear optimization.)
    The discretization order of the finite difference pre-processing is a hyperparameter of SINDy. The first order accurate finite difference (here referred to as FD-1) results in a point-wise approximation of,
    \begin{equation}
        \frac{x_{n+1}-x_n}{\Delta t} = \Xi \phi(x_n).
    \end{equation}
    Note that when rearranged, this is equivalent to the forward Euler integrator:
    \begin{equation}
        x_{n+1} = x_n + \Delta t\,\Xi \phi(x_n).
    \end{equation}
    Higher-order finite difference stencils can also be used to increase the accuracy of the time derivative approximation. The second-order stencil (FD-2) (analogous, but not equivalent, to Midpoint) can be expressed as
    \begin{equation}
        \frac{1}{2\Delta t}\left(x_{n+1}-x_{n-1}\right) = \Xi \phi(x_n).
    \end{equation}
    and the fourth-order stencil (FD-4) (analogous, but not equivalent, to RK4) can be expressed as,
    \begin{equation}
        \frac{1}{\Delta t}\left(
        -\frac{1}{12}x_{n+2}
        +\frac{2}{3}x_{n+1}
        -\frac{2}{3}x_{n-1}
        +\frac{1}{12}x_{n-2}
        \right) = \Xi \phi(x_n).
    \end{equation}

\begin{figure}[t]
    \centering
    \includegraphics[scale=0.7]{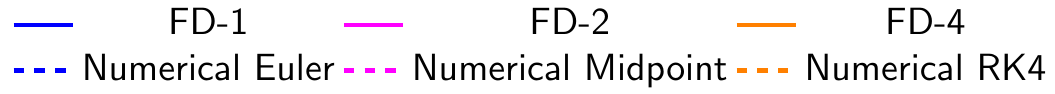} 
    \subfloat[Harmonic oscillator: $\Delta t$ = 0.1] {{ \includegraphics[width=0.3\linewidth]{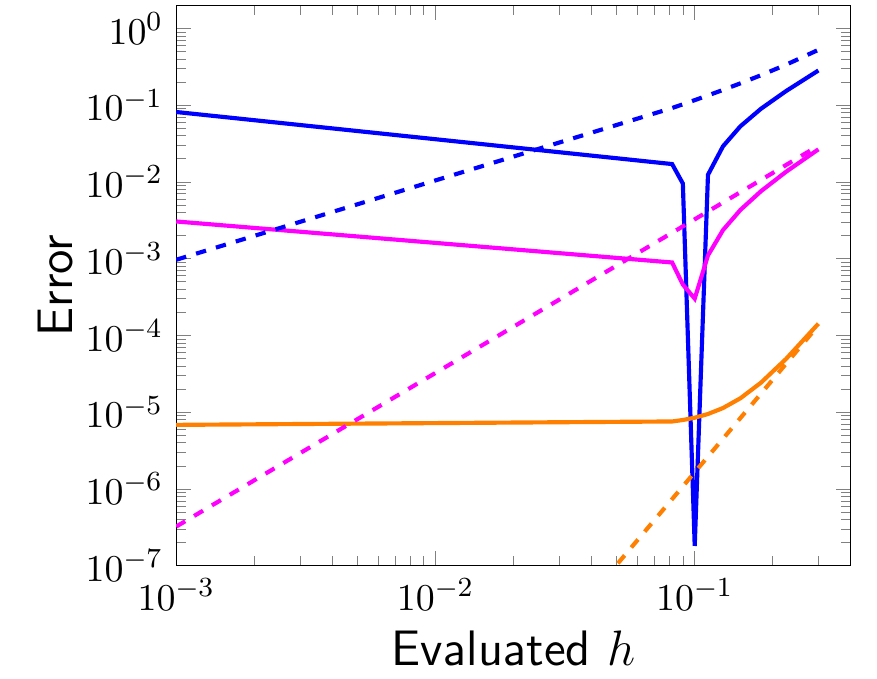}
    }} 
    \subfloat[Pendulum: $\Delta t$ = 0.1] {{ \includegraphics[width=0.3\linewidth]{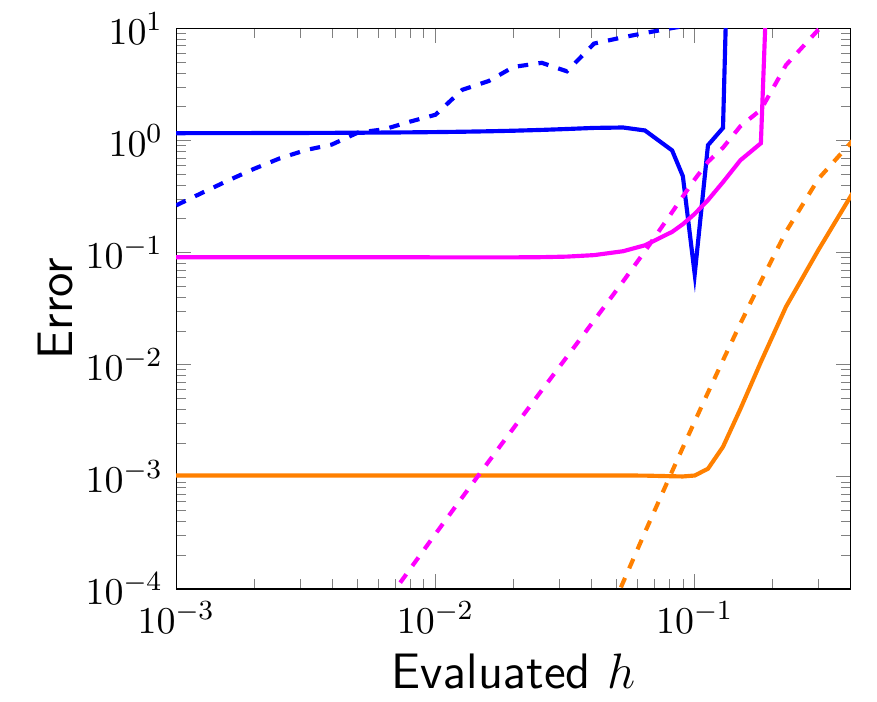}
    }} 
    \subfloat[Pendulum: $\Delta t$ = 0.05] {{ \includegraphics[width=0.3\linewidth]{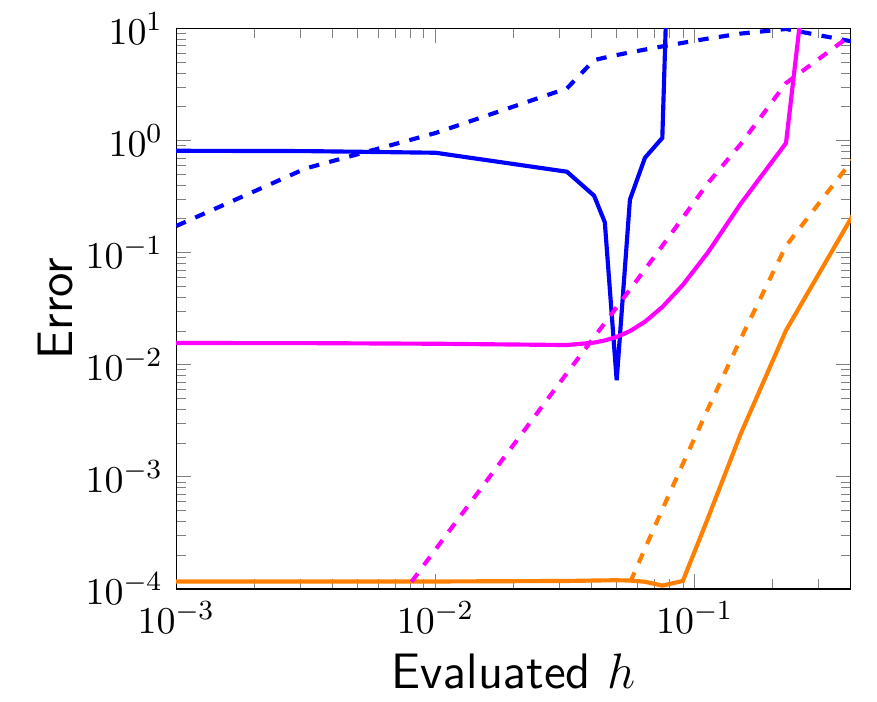}
    }} 
    \caption{\textbf{SINDy method: additional convergence tests.}  Instead of training from dynamical data with an ODE-Net model, SINDy is used instead to learn a sparse polynomial basis. The accuracy of the time derivative approximation is altered by using a different order of central finite difference stencils. Here, FD-1 is the first-order two-point stencil, FD-2 is the second-order three-point stencil, and FD-4 is the fourth-order five-point stencil. In this example, FD-1 is analogous to Euler-Net, FD-2 is analogous to Midpoint-Net, and FD-4 is analogous to RK4-Net. In (a), the convergence test method is shown on the harmonic oscillator, where the convergence test is replicated by training via the SINDy method. In (b) and (c), the convergence test is applied when training on the non-linear pendulum via the SINDy method. 
    We observe similar results as when training using an ODE-Net: the lower-order approximation (FD-1) overfits to the $\Delta t$ in the training data, failing the convergence test; and the higher-order approximation (FD-4) passes the convergence test.
    }
    \label{fig:sindy_additional convergence}
\end{figure}

\paragraph{Training Setup.} We look at the harmonic oscillator described in~\eref{eq:eqn_harmonic_oscillator}, and the non-linear pendulum (described in~\eref{eq:non_linear_pendulum} in~\nameref{sec:ds_systems}). We use the PySINDy implementation~\cite{desilva2020, Kaptanoglu2022} to train models on these trajectories. We train three different SINDy models on the trajectories, altering the finite difference (FD) approximation order of accuracy: FD-1 is a first-order two-point stencil (analogous to Euler-Net), FD-2 is a second-order three-point stencil (analogous to Midpoint-Net), and FD-4 is a fourth-order five-point stencil (analogous to RK4-Net). The SINDy model is plugged into the convergence test as $F$, such that the same Runge-Kutta integrators are used for trajectory prediction. For the harmonic oscillator, the training data is spaced apart by $\Delta t = 0.1$, while for the non-linear pendulum, we look at examples where the training data is spaced apart by $\Delta t = 0.05$ and $\Delta t = 0.1$.

\paragraph{Results.} We run our convergence test on the different SINDy models. The results of our method are shown in~\fref{fig:sindy_additional convergence}. In each case, the FD-1 model has low error when $h = \Delta t$ (i.e., evaluated at the same time spacing as the training data), but it has high error when evaluated at all other $h$, and especially smaller values of $h$. Thus, it does not pass the convergence test, and it has not learned a meaningfully continuous dynamics. 
The FD-1 model shows a sharp dip because it is overfit to forward Euler. In this case, the stencil and integrator correspond to the exact same algebraic structure.
Similar to when the models where trained via ODE-Nets, we see that FD-1 has overfit to the temporal discretization. In contrast, the error during inference time of the FD-4 model steadily decreases when it is evaluated at lower $h$, eventually converging to a fixed basal level. It has passed the convergence test, and it has learned a meaningfully-continuous model.

%% file: _s5_conclusions_comm_physics.tex
\section*{Conclusion}

One of the great challenges in scientific ML is to learn continuous dynamics for physical systems---either ``the'' underlying continuous dynamics, or ``a'' continuous dynamics that leads to good predictive results for the spatial/temporal regime of interest for the ML model---such that the learned ML model can be trusted to give accurate and reliable results.
ML models are trained on discrete points, and typical ML training/testing methodologies are not aware of the continuity properties of the underlying problem from which the data are generated.
Here, we have developed a ContinuousNet methodology, and we showed that convergence (an important criteria used in numerical analysis) can be used for selecting models that have a strong inductive bias towards learning meaningfully continuous dynamics. 
Standard ODE-Net approaches, as well as common SINDy methods, both popular in recent years within the ML community, often do not pass this convergence test.
In contrast, models that pass this convergence test have favorable properties. 
For instance, models that learned underlying continuous dynamics can be evaluated at lower or higher resolutions. 
Our results suggest that principled numerical analysis methods can be coupled with existing ML training/testing methodologies to deliver upon the promise of scientific ML more generally.

Many more concrete directions are of course raised by our methodology.
One direction has to do with developing analogous tests to be used for less well-posed dynamical systems.
Such systems are of interest in scientific ML, and such tests will be of greatest interest when one needs to obtain ``the'' correct underlying continuous solution (e.g., to identify correctly qualitative long-term behavior~\cite{ker_disc_cont_92}), rather than ``a'' continuous solution, which is often sufficient for ML prediction tasks.
Another direction has to do with whether we can develop analogous tests appropriate for adaptive time-stepping methods, symplectic integrators, and other commonly used numerical simulation methods such as those for optimal control problems using NNs and associated Hamilton-Jacobi partial differential equations~\cite{NGW21,DLM20}.
Work subsequent to the posting of the initial technical report version of this paper has addressed the continuous-discrete equivalence question for learning operators~\cite{BBRMMA23_TR,RMRRBx23_TR}, and likely our ContinuousNet methodology provides a way to operationalize that in practice.
A final direction has to do with whether one can obtain strong theoretical results, e.g., ML-style generalization bounds, to guide the use of methods such as these.
Recent theoretical and empirical results suggest that this will be challenging~\cite{MM19_HTSR_ICML,MM20_SDM,MM20a_trends_NatComm,MM21a_simpsons_TR,liam_generaliz_bounds_ICML22}, at least when using traditional approaches to ML-style generalization bounds.
Our success in combining
principled numerical analysis methods with existing ML methodologies also leads one to wonder whether we can use
a posteriori error bound analysis methods to develop practically useful a posteriori generalization bounds for problems such as those we have considered.

%% file: _s3_problem_overview_comm_physics.tex
\section*{Methods}
\label{sec:methods}

\label{sec:methods}

The basic problem of numerical analysis is to solve problems from continuous mathematics using a discrete computer.
The area has a rich history for describing the consistency and convergence behavior of numerical methods for approximating continuous functions~\cite{moin2010fundamentals, leveque1992numerical}. Here, we expand on the methods we used in~\nameref{sec:results}.

\subsection*{Criteria of Classical Numerical Analysis}
\label{subsec:criteria_numerical_analysis}

Given an initial value $x(0) = x_0$, we can discretize Eq.~\ref{eq:de} along the node points $t_n = n \Delta t$ for $n=0,1,\dots,N$ by evaluating the following integral equation:
\begin{equation}
x_{n+1} = x_{n} + \int_{t_n}^{t_n+\Delta t} F(x(s)) \, \mathrm{ds},
\label{eq:numerical_integration}
\end{equation}
where $x_n = x(t_n)$, and $\Delta t$ is the discrete timestep. Typically, we are not able to compute an analytic solution for the integral, and thus we rely on numerical schemes to approximate $ \int_{t_n}^{t_n + \Delta t} F(x(s)) \, \mathrm{ds}$.

There are many different types of numerical integration schemes to approximate the integral in Eq.~\ref{eq:numerical_integration}.
These have different trade-offs between computational efficiency and accuracy. 
One such scheme, the forward Euler discretization, can be written as:
\begin{equation}
    x_{n+1} = x_{n} + \Delta t \, F(x_n).
\label{eq:forward_euler}
\end{equation}
This is a first-order one-step method, where the global error (the error over all of the timesteps) is proportional to the step size, i.e., $\mathcal{O}(\Delta t)$, meaning that the error gets smaller as $\Delta t$ decreases. 
There are also higher-order integration schemes. 
One popular higher-order scheme is the Runge-Kutta 4 (RK4) discretization, which takes the following form:
\begin{equation}
\begin{split}
i_{1} & = F\left(x_n\right) \\
i_{2} & = F\left(x_n + \frac{\Delta t}{2} \cdot i_{1}\right) \\
i_{3} & = F\left(x_n + \frac{\Delta t}{2} \cdot i_{2}\right) \\
i_{4} & = F\left(x_n + \Delta t \cdot i_{3} \right) \\
x_{n+1} & = x_n + \frac{1}{6} \Delta t \left(i_{1} + 2i_{2} + 2i_{3} + i_{4}\right).
\end{split}
\label{eq:rk4}
\end{equation}
Here, the global error is proportional to the step size to the fourth power, i.e., $O(\Delta t^4)$; and thus as $\Delta t$ gets smaller, the error gets smaller much more quickly than with the forward Euler scheme. In general, the global error can be written as $O(\Delta t^p)$, where $p$ denotes the order of accuracy.

%

Classical numerical integration typically starts by assuming that there exists a true underlying continuous-time system, which is then replaced by a discrete-time problem whose solution approximates that of the continuous problem. 
However, discretizing the problem introduces an error, and concepts such as stability, convergence, and consistency can be used to quantify the error of the discrete solution~\cite{dahlquist1956convergence, arnold2015stability, kirby2008need}.

In the following, we describe the error bounds in the traditional scientific computing context where the system dynamics are known exactly, in which case the only approximation error comes from numerical integration in time.
We specifically focus on numerical convergence because this will give us a mechanism to analyze ML models. However, note that stability and consistency are also of interest~\cite{leveque1992numerical}.
Let $x(t_n)$ denote the true solution of a dynamical system of interest; and
let $\bar{x}^{\Delta t}_{n}$ denote a numerical solution after $n$ steps with step size $\Delta t$. {We use $N$ to denote the maximum number of time steps such that $T = t_N = N \Delta t$ is the final time. Decreasing $\Delta t$ requires increasing $N$ (the number steps taken to arrive at $T$), and vice versa. Then, $x(T)=x(t_N)$ is the true solution at the final time, while $\bar{x}^{\Delta t}_{N}$ is the numerical solution at the final time. }
Convergence quantifies the global error (the cumulative error of all iterations) of a numerical algorithm. 
\begin{definition}[Convergent Numerical Approximation.]\label{def1}
	A numerical one-step method for solving $\frac{\mathrm{d}x(t)}{\mathrm{d}t} = F(x(t))$, with initial condition $x(0)=x_0$, is said to be convergent if and only if the error tends to zero as $\Delta t$ goes to zero:
	\begin{eqnarray}
\lim_{\Delta t \rightarrow 0} 
    \,\,\, \left\| x(t_N) -\bar{x}^{\Delta t}_{N}\right\|_2 = 0. 
	\label{eq:convergence1}
	\end{eqnarray}
\end{definition}

\noindent
Of course, in numerical practice, the error does not converge to zero.
Instead, it levels off at some base level determined typically by the level of numerical precision used to describe the data, as observed in \ref{fig:convergence_test_schematic}\protect\subref{subfig:harm_osc_inference_rk4}.

The specific metric for quantifying the approximation error across the sequence is somewhat arbitrary.
Moreover, there is the problem that the numerical method can potentially converge to the wrong solution~\cite{thompson1992numerical}.
Thus, to ensure that a numerical method is not only convergent but also consistent, one can use the mean error,
$$
\lim_{\Delta t \rightarrow 0} 
\,\,\, \frac{1}{N}\sum_{n=0}^N \left\| x(t_n) -\bar{x}^{\Delta t}_{n}\right\|_2 = 0,$$
or the maximum error across all $N$ points in time,
$$
\label{eq:global_error_all_points}
\lim_{\Delta t \rightarrow 0} 
\,\,\,\,\, \max_{n=1,2,\dots,N} \,\,\, \left\| x(t_n) -\bar{x}^{\Delta t}_{n} \right\|_2 = 0.
$$
%
%
If, as the step size $\Delta t$ decreases, the largest absolute error between the numerical solution $\bar{x}^{\Delta t}_{n}$ and the exact solution $x(t_n)$ also decreases, then the numerical approximation converges towards the solution of the continuous system.
In the limit of $\Delta t \rightarrow 0$, the numerical solution converges to the exact solution and the error converges to zero, or to some base level determined by machine precision and numerical round-off noise.
%

This convergence criteria is also a test for continuity in the solution: as $\Delta t \rightarrow 0$, the time interval between adjacent numerical solutions (e.g., at $t_{n}$, $x_{n}$, and at $t_{n+1}$, $x_{n + 1}$) also decreases towards zero. Thus, the numerical solution collapses onto a continuous solution as $\Delta t \rightarrow 0$. 

\begin{remark}
\normalfont
	Validation of a new integration method involves multiple stages. Consistency, convergence, and stability can be theoretically proven for a rather small class of ODEs (typically only linear ODEs). 
	Thus, the method will be evaluated empirically with a real implementation on a problem of interest.
	In practice, a convergence test is used, where the numerical integration scheme is used to predict trajectories for a range of $\Delta t$ and compared to an analytical solution or to an overrefined solution.
	The errors are verified to approach zero at the correct rate, at least until they flatten out at some base level.
	The combination of theoretical proof of consistency, stability, and convergence on simple systems such as linear ODEs, combined with the emprical demonstration of convergence on ODEs of interest, is typically viewed as sufficient to vet the method.
	Emprirical convergence tests are a standard integration test method for scientific programs, e.g., they are regularly run to automatically catch bugs.
\end{remark}


\subsection*{A Convergence Test for ODE-Nets}
\label{subsec:method_convergence}

We now describe a convergence test, based on the discussed convergence criteria, to validate properties of an ODE-Net solution. %
The fact that ODE-Nets are embedded in a numerical integration scheme enables us to use convergence analysis methods that are well-known for studying classical numerical analysis problems. 
To start, we know that the numerical integrator itself will be convergent if it is given the true $f$, but we do not know if the ODE-Net will be convergent when an approximate ML model $\mathcal{N}$ is used to approximate $f$.
The convergence test is used to determine whether an ODE-Net has learned a meaningfully continuous model for the underlying problem of~interest.

Suppose that we are given an ODE-Net $\mathcal{N}$ that is trained with a numerical integration scheme (such as Euler or RK4) from $t_n$ to $t_{n+1}$ with stepsize $\Delta t$:
\begin{eqnarray}
    \bar{x}_{n+1} = \text{ODESolve} [ \mathcal{N}(x_n; \theta),\,\, \text{start}=t_n, \,\, \text{end}=t_{n+1},\,\, \text{step}=\Delta t,\,\, \texttt{scheme} ]  . 
    \label{eq:full_odenet}
\end{eqnarray}
Following Definition~\ref{def1}, we compute the global error of the ODE-Net $\mathcal{N}$ as it approaches some fixed value $b$ as the time step $h$ goes to zero:
\begin{equation}
\lim_{h \rightarrow 0} 
 \,\,\, \left\| x_N -\bar{x}^{h}_{N} \right\|_2 = b.
	\label{eq:convergence2}
\end{equation}
Here too, in the ML setting, the error does not necessarily converge to zero. 
This is analogous to the classical numerical analysis setting, where the numerical analysis test typically converges to a non-zero value determined by the numerical round-off~error (e.g., see the floor in Figure~\ref{subfig:harm_osc_inference_rk4}).  
 Unlike in the classical numerical analysis setting, even in the absence of numerical errors the $b$ of an ML model will be greater than zero. 
For an ODE-Net, the numerical value of $b$ depends on the model architecture, integration method, the optimizer, and the noise properties of the data~\cite{bottou2007}.
The value of $b$ will elucidate the convergence properties of the trained ODE-Net. 

Computing an error metric in the ML setting requires additional consideration because we do not necessarily have access to the underlying exact solution at arbitrary points in time. 
Instead, we are restricted to the information that is provided by a given validation set of discrete data points $\mathcal{T} = \left\{x_0, \, x_1, \, x_2, \, \dots, \, x_N\right\}$, where each point is spaced apart by the $\Delta t$ between observations. 
A naive metric to compute the global error in this setting is simply to consider the $2$-norm between the end point of the validation trajectory and the predicted value:
\begin{equation}
    \mathrm{Error}(h) = \left\| x_N -\bar{x}^{h}_{N} \right\|_2.
    \label{eq:conv_test_error1}
\end{equation}
However, this metric is susceptible to noise and edge cases.
Computing the error over all points $\mathcal{T}$ is difficult using~\eqref{eq:conv_test_error} because the inferred trajectory has a different number of points than the validation trajectory.
To mitigate this issue, we suggest to compute the global error on a subset of points $\mathcal{S}$ from the validation/test trajectories, which is a set of indexes into the original dataset $\mathcal{T}$; e.g., $\mathcal{S}=\left\{0, 9, 18\dots\right\}$ for every 9 points spaced by $9\Delta t$. This allows for inferring the trajectory of $h$ computing the error over the subset as follows:
\begin{equation}
    \mathrm{Error}(h) = 
      {\frac{1}{|\mathcal{S}|} \sum_{n\in \mathcal{S}} \left\| x_{n} -  \bar{x}^{h}_{k} \right\|_2},
    \label{eq:conv_test_error}
\end{equation}
where $k$ is the index into the inferred trajectory corresponding to the index into the validation trajectory $n$ such that $\bar{x}^{k}_{k}$ is the point that lines up at the same time as $x_{n}$.
Note that we can only use certain timesteps $h$ during inference because the solution points must align perfectly with those in the subset trajectory.

Given this setup, we use the term ContinuousNet to refer to an ODE-Net model that also exhibits these convergence properties, as per Definition~\ref{def:continuousnet}.
To evaluate this property, we can apply the same convergence test procedure used in traditional numerical analysis and scientific computing, but with the modifications necessary for it work on training data.
Further, it is necessary to apply a weaker heuristic to judge convergence because there will be residual optimization error at $Error(\Delta t)$, as per \eref{eq:convergent_criterion}.
The procedure is as follows.

Given the learned ODE-Net, we first infer a validation/test trajectory on the original stepsize $\Delta t$ and evaluate the global error, $Error(\Delta t)$. This is the standard procedure for evaluating a discrete model, and this error value informs its accuracy at inferring discrete points in a sequence.
Then, to further evaluate whether the model is convergent and continuous, we consider a range of $h$ values which are both smaller and larger than the step size $\Delta t$ used during training. For example, if the ODE-Net was trained on $\Delta t = 0.1$, we evaluate the ODE-Net on the validation/test trajectory for $h \in [10^{-3},..,10^{1}]$.
Specifically, for a given set of inference timesteps $\{h_1,h_2,\dots,h_p\}$, our proposed convergence test iterates over the elements $h_i$, and executes the following two steps:
\begin{enumerate}
\item Evaluate the pre-trained ODE-Net using~\eref{eq:full_odenet} on the time interval $[t_0, t_{T}]$, using step size~$h_i$. 
\item Calculate the error between the ODE-Net solution $\bar{x}^{h_i}_{n}$ and points in the test data, $Error(h_i)$.
\end{enumerate}
Algorithms~\ref{alg:forward} and~\ref{alg:forward2} in~\nameref{sxn:app_algorithms} summarize this procedure. As with many other numerical convergence tests, our proposed algorithm is subject to a heuristic threshold. In practice, we observe that the difference between a model that passes our test condition and one that does not is pronounced. A non-continuous model results in an error that is orders of magnitude larger, as compared to a continuous model, when $h$ is taken smaller than $\Delta t$ of the training data.
In other words, our convergence test performs a form of model selection by selecting for models that learn inductive biases towards meaningfully continuous~dynamics. In the following, we will demonstrate why convergence is an important consideration for ML model~selection.\footnote{We also discuss the two other criterion of classical numerical analysis, consistency and stability, in~\nameref{sxn:app_algorithms}.} 

\begin{remark}
\normalfont
In practice, we can also evaluate our convergence test with different starting points $x_{0}$ and the respective final time step, $x_{N}$, and then average the error across the different runs. We highly recommend this to ensure that the same  behavior occurs irrespective of start and end point.
\end{remark}


\subsection*{Error Analysis in an Idealized Learning Setting}
\label{subsec:theory_error_analysis_idealized}

\begin{figure*}[!ht]
	\centering
    \includegraphics[scale=0.7]{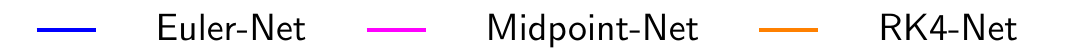} 
	\subfloat[Analytical global error with $\varepsilon = 0$] {{ \includegraphics[width=0.33\linewidth]{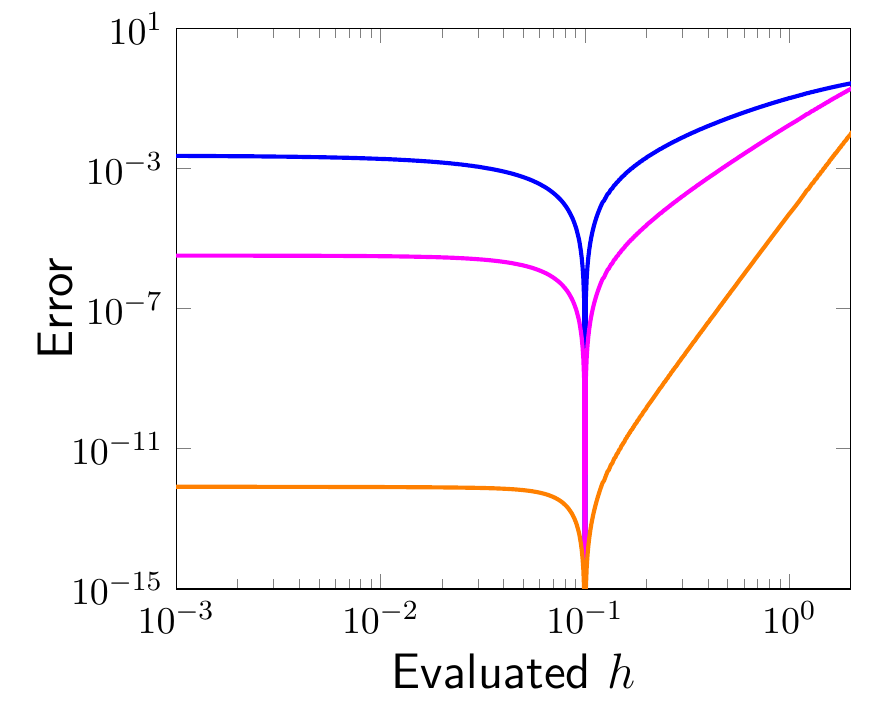}
	\label{subfig:analytical-0}
	}} 
	\subfloat[Analytical global error with $\varepsilon = 10^{-5}$] {{ \includegraphics[width=0.33\linewidth]{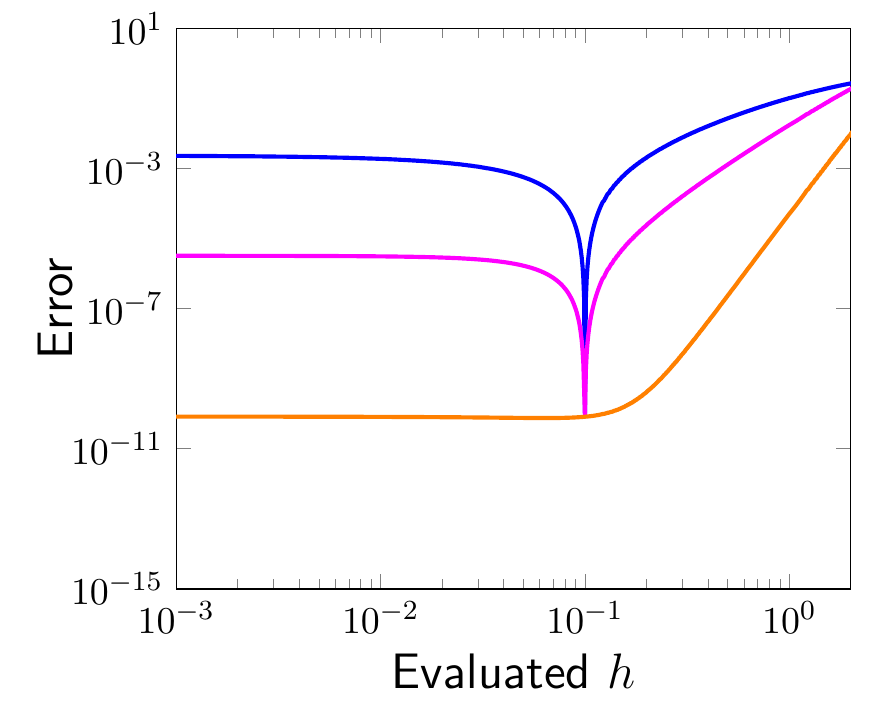}
	\label{subfig:analytical-1e-5}
	}}  \
	\caption{\textbf{Illustration of our convergence test for linear ODEs.} 
		\footnotesize The analytically derived global error estimate, $Error(h)$, using the ideally learned linear ODE. Our results replicate the empirically performed convergence tests. (a) All methods achieve zero global error at $h=\Delta t$, but converge to finite values as $h\rightarrow 0$. (b) We introduce the $\varepsilon$ error parameter. idpoint and RK4 achieve a finite error of comparable magnitudes at $h=\Delta t$. Overfitting is not observed for RK4 with $\varepsilon=10^{5}$ because the $|\lambda-w|$ error is less significant than $|\varepsilon|$.}

	\label{fig:linear_ode_convergence_test}
\end{figure*}

We provide a theoretical framework for the convergence test, analyzing the discretization error of one-step numerical integration schemes in an idealized setting. Specifically, we consider the problem of learning the simple scalar linear ODE,
\begin{equation}
\frac{\mathrm{d}x(t)}{\mathrm{d}t} = \lambda \, x(t),
\end{equation}
where $\lambda \in \mathbb{R}$ denotes a scalar parameter, and $x \in \mathbb{R}$ denotes the state at a given point in time. It is well known that the function $x(t)=e^{\lambda t}x_0$ is a solution of this system, given the initial condition $x(t)=x_0 \in \mathbb{R}$~\cite{hirsch2012differential}.
In the following, we assume a similar setting as before: we are given a set of discrete data points $\mathcal{D} =\left\{x_0, \, x_1, \, x_2, \, \dots, \, x_N\right\}$ produced by the linear system and spaced by $\Delta t$.
Our aim is to learn a scalar ODE-Net model,
\begin{equation}
\frac{\mathrm{d}x(t)}{\mathrm{d}t} = w \, x(t),
\end{equation}
parameterized by the learnable weight parameter $w$. Following the ODE-Net process, we discretize the model with a numerical integration scheme and then optimize the squared error loss,
\begin{equation}
\label{eq:analytical_optimization_maintext}
    \min_{w} \sum_{x\in \mathcal{D}} \left(x_{n+1}-
    \text{ODESolve}\left[ wx_n,\,\, \text{step}=\Delta t,\,\, \texttt{scheme} \right]\right)^2.
\end{equation}
We assume that the data are noise-free and can therefore be represented by its analytical solution, $x_{n+1}=e^{\lambda \Delta t}x_n$.
When the loss is optimized, the time-discretization step introduces its own unique source of error into the learning process, one that is independent of noise, numerical error, or optimization error. The error stems from the fact that any one-step consistent numerical integration scheme, when applied to a linear ODE, will result in a truncated Taylor series expansion with $p$ terms, where $p$ is the accuracy order of the scheme~\cite{leveque1992numerical}. Thus, the ML model cannot recover the exact parameters of the underlying ODE.
The following lemma makes this issue explicit.

\begin{lemma}
\label{thm:polynomial_root} In absence of any other optimization errors, a scalar ODE-Net can at best obtain a weight parameter $w$ by minimizing Eq.~\eqref{eq:analytical_optimization_maintext} that, for certain timesteps and integrators, satisfies the following polynomial equation,
\begin{equation}
\label{eq:polynomial_roots_main}
    \exists p,\,\Delta t,\,w \quad s.t.\quad \sum_{i=0}^p \frac{\Delta t^i}{i!} {w}^i = e^{\lambda \Delta t}.
\end{equation}
\end{lemma}
The proof is given in \nameref{sec:appendix_proof_1}. For finite $\Delta t$, this equation clearly satisfies $w\neq \lambda$ if $p\ll \infty$. There are situations where this equation can be solved for values of $w$ that will set the loss in Eq.~\eqref{eq:analytical_optimization_maintext} to zero for all possible data points $x_n$. In the limit as $\Delta t \rightarrow 0$, there is always a solution at $w=\lambda$ for any $p$. Moreover, for practical settings there is at least one root when $p$ is odd for any $\Delta t$; when $p=2$ (RK2) and $\Delta t\leq\log(2)/|\lambda|$; or when $p=4$ (RK4) and $\Delta t<1.307/|\lambda|$. This equation can be used to find analytical expressions for $w$ for simple integrators; see~\nameref{sec:appendix_example_equations}.

From this result, we can characterize the difference between the ML model and the target ODE.
In addition, in practice, it is (almost always) only possible to learn a perturbed version, $\tilde{w}$, of ${w}$ due to noise, limited numerical precision, and optimization errors. Let $\varepsilon$ denote an additive perturbation away from the the optimum of the minimization problem for an observed model due to these sources of errors, $\tilde{w}=w+\varepsilon$.
The following theorem bounds the error between the ML and ODE model, due to the overfitting of Eq.~\eqref{eq:polynomial_roots_main} in presence of additive error sources. 
\begin{theorem}
\label{thm:wminuslambda}
  If optimal weight parameter can be found by Eq.~\eqref{eq:polynomial_roots_main}, the approximation error introduced by scalar ODE-Net is bounded by
$|{w}-\lambda|\leq c \Delta t^p$, where $c$ is a constant proportional to the Lipschitz continuity constants of $\lambda x$ and $w x$. In the presence of additive numerical error $\varepsilon$, the bound is,
\begin{equation}
\label{eq:error_bound_eps}
|\tilde{w}-\lambda | \leq |{w}-\lambda| + |\varepsilon| \leq c \Delta t^p + |\varepsilon|.
\end{equation}
\end{theorem}
The proof is given in \nameref{sec:appendix_proof_2}.
This bound shows that the user needs to increase the order of accuracy of the training scheme $p$ in order to reduce the error between the learned parameter and the true ODE parameter.

In practice, we can only measure $\mathrm{Error}(h)$ using a set of data points. Using the above results, we can analyze the expected behavior of the convergence test by bounding the global error using Eq.~\eqref{eq:error_bound_eps}. \fref{fig:linear_ode_convergence_test} plots the global error bound given concrete values of $\lambda, \Delta t$, and $\varepsilon$. As can be seen, the theoretically derived global error for the scalar ODE exhibits the same behavior as the empirical applications of the convergence test.
We can use the global error to further derive expected bounds on the the key points of the convergence test.

\begin{corollary}
\label{thm:error_dt}
When a scalar ODE-Net is evaluated with the timestep that was used for training, the leading term of the global error is proportional to the optimization error $\varepsilon$ for a $k\propto T |x_0|$:
\begin{align}
    \mathrm{Error}(\Delta t) & = k |\varepsilon| + \mathcal{O}(\Delta t |w| |\varepsilon| + \Delta t |\varepsilon|^2).
\end{align}
\end{corollary}
The proof is given in \nameref{sec:appendix_proof_error}.
 The $c\Delta t^p$ error term in Eq.~\eqref{eq:error_bound_eps} between $w$ and $\lambda$ is cancelled out, and the observed value is smaller than $|\tilde{w}-\lambda|$. Therefore, the global error only observes the difference between $w$ and $\tilde{w}$, resulting from the model optimization error. Note that this value can become very small as the optimization error decreases. By applying the convergence test, we are able to extract an estimate of $|\tilde{w}-\lambda|$, as given in the following.
\begin{corollary}
\label{thm:error_h_0}
In the limit of decreasing the timestep size during inference, the global error approaches a constant factor ($b$) based on the bound in Eq.~\eqref{eq:error_bound_eps}. It approaches at a rate of $h^q$, where $q$ is the accuracy order of the ODE integration scheme used at inference time:
\begin{align}
    \lim_{h\rightarrow 0}\mathrm{Error}(h) & = |\tilde{w} - \lambda | + \mathcal{O}(h^q) \leq k|\varepsilon| + k c \Delta t^p + \mathcal{O}(h^q).
\end{align}
\end{corollary}
The proof is given in \nameref{sec:appendix_proof_error}.
Given these bounds, we can see how using Eq.~\eqref{eq:convergent_criterion} as a threshold yields $(b-\mathrm{Error}(\Delta t)) \propto c \Delta t^p$.
Therefore, the comparison between $b$ and $\mathrm{Error}(\Delta t)$ allows for the quantitative estimation for the magnitude of the term $c \Delta t^p$, which describes the error that is induced by the numerical discretization scheme used for training.

In summary, our analysis shows that there are two types of errors in the process of learning the dynamics of a linear scalar ODE using ODE-Nets. These errors can be measured using the data by evaluating $\mathrm{Error}(\Delta t)$ and $\lim_{h\rightarrow 0}\mathrm{Error}(h)$.
This illustrates the power of our proposed convergence criterion Eq.~\eqref{eq:convergent_criterion}. Moreover, even if $\varepsilon$ is small, it is required to increase the order of accuracy of the training scheme in order to further decrease the error between $\tilde{w}$ and $\lambda$.

\color{black}

%% file: _s6_acknowledgements.tex
\paragraph{Acknowledgements.}
We would like to thank Annan Yu and Krishna Harsha Reddy Kothapalli for valuable discussions and feedback. Moreover, we would like to thank all the reviewers for their helpful and constructive feedback.
ASK was supported by Laboratory Directed Research and Development (LDRD) funding under Contract
Number DE-AC02-05CH11231 at LBNL and the Alvarez Fellowship in the Computational Research
Division at LBNL.
MWM would like to acknowledge the DOE, NSF, and ONR for providing partial support of this work.  
NBE would like to acknowledge support from NSF (DMS-2319621), DOE (AC02-05CH11231), and NERSC (DE-AC02-05CH11231). Our conclusions do not necessarily reflect the position or the policy of our sponsors, and no official endorsement should be inferred.

%% file: _s7_supplementary.tex
\appendix 

\section*{Supplementary Note 1: Details about considered dynamical systems}
\label{sec:ds_systems}

\paragraph{Non-linear pendulum.} 
The non-linear pendulum is similar to the harmonic oscillator (without the small-angle approximation) and is governed by a second-order differential equation. It can be written as a first-order equation with two degree-of-freedom:
\begin{equation}
	\frac{\mathrm{d} \theta}{\mathrm{d}t}  = v; \quad \quad
	\frac{\mathrm{d} v}{\mathrm{d}t}  = -\omega_{0}^2 \,\, sin \theta,
	\label{eq:non_linear_pendulum}
\end{equation}
where $\theta$ is the angle of the pendulum, $v=\mathrm{d}\theta/\mathrm{d}t$ is the rate-of-change of angle, and $\omega_0$ the resonance frequency based on gravity and mass, $\omega_0=\sqrt{g/m}$. The analytical solution can be found in \cite{belendez2007exact}.

\paragraph{Lotka-Volterra equations.} 
The Lotka-Volterra predator-prey model is a set of one-dimensional non-linear differential equations. 
The model describes the dynamics of a biological system where predators and prey interact. 
The population in each species, denoted by $x$ and $y$, is modeled over time according to the equations,
\begin{equation}
	\frac{\mathrm{d}x}{\mathrm{d}t} = ax - bxy ; \quad \quad
	\frac{\mathrm{d}y}{\mathrm{d}t} = cxy - dy,
\end{equation}
where $a, b, c$ and $d$ are population dynamics coefficients. Training data was obtained by numerical simulation with a very tiny timestep, $h=\Delta t/1000$, using RK4 to mitigate numerical error artifacts in the training data.

\paragraph{Cartesian pendulum.} 
This is the non-linear $\theta$ pendulum in Cartesian space, now represented in x-y coordinates. This increases the degrees of freedom in the system to four. A representation as an ODE requires force to satisfy the constraint that the pendulum mass at a fixed distance away from its center, $x^2+y^2=L$. Differentiating the constraint twice yields the expression $f = (v_x^2+ v_y^2) + g\,(-y/L)$.
With this formula, the coupled equations are,
\begin{align*}
	\frac{\mathrm{d}x}{\mathrm{d}t} &= v_x \quad &
	\frac{\mathrm{d}y}{\mathrm{d}t} &= v_y \\
	\frac{\mathrm{d}v_x}{\mathrm{d}t} &= f\frac{x}{L} \quad &
	\frac{\mathrm{d}v_y}{\mathrm{d}t} &= f\frac{y}{L} + g.
\end{align*}
The Cartesian pendulum is an example of a \textit{stiff} ODE, where numerically solving such a system can easily be numerically unstable without special consideration. We generated training data using the analytical solution to in the pendulum in $\theta(t)$ from \cite{belendez2007exact}.

\paragraph{Double gyre fluid flow.} 
The double gyre fluid flow 
is a benchmark model for learning temporal dynamics from fluid flow field snapshots. The system dynamics is described by unsteady Stokes equations, and has a closed form analytical solution using the stream function,
\begin{equation}
	\begin{split}
		\phi(x, y, t) &= A \,\, \sin(\pi f(x, t))  \,\, \sin(\pi y),  \\
		f(x,t) &= a(t)x^2 + b(t)x,  \\
		a(t) &= \epsilon  \,\, \sin(\omega t),  \\
		b(t) &= 1 - 2 \epsilon  \,\, \sin(\omega t).
	\end{split}
\end{equation}
The velocity field is computed from the stream function,
\begin{equation}
	\begin{split}
		u(x,y,t) = -\frac{\partial \phi}{\partial y} = - \pi A  \,\, \sin(\pi f(x)) \cos(\pi y), \\
		v(x,y,t) = \frac{\partial \phi}{\partial x} = \pi A  \,\, \cos (\pi f(x)) \sin(\pi y) \frac{df}{dx}.
	\end{split}
\end{equation}

\section*{Supplementary Note 2: Model architecture details}
\label{sxn:app_architecture}
The base model for the harmonic oscillator is a linear model with only four parameters. In this case, the Euler-Net is a linear model, but the RK4-Net is a non-linear model.
The base model for the nonlinear pendulum and Lotka-Voltera equations is a shallow neural network with a hidden dimension of 50. 
The Cartesian pendulum uses a shallow neural network with a hidden dimension of 2000.
The Double gyre fluid flow problem uses a convolutional autoencoder with 4 layers in the encoder and decoder to learn a latent space of 5 dimensions, with a shallow neural network to learn an ODE on the latent space. 
All problem settings use the Adam optimizer with weight decay.

\section*{Supplementary Note 3: Additional convergence test examples using ODE-Nets}
\label{sec:appendix_additional_examples_convergence}

\begin{figure}[!ht]
    \centering
    \includegraphics[scale=0.7]{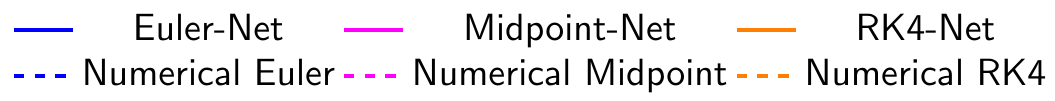} 
    \subfloat[Trained $\Delta t$ = 0.01] {{ \includegraphics[width=0.3\linewidth]{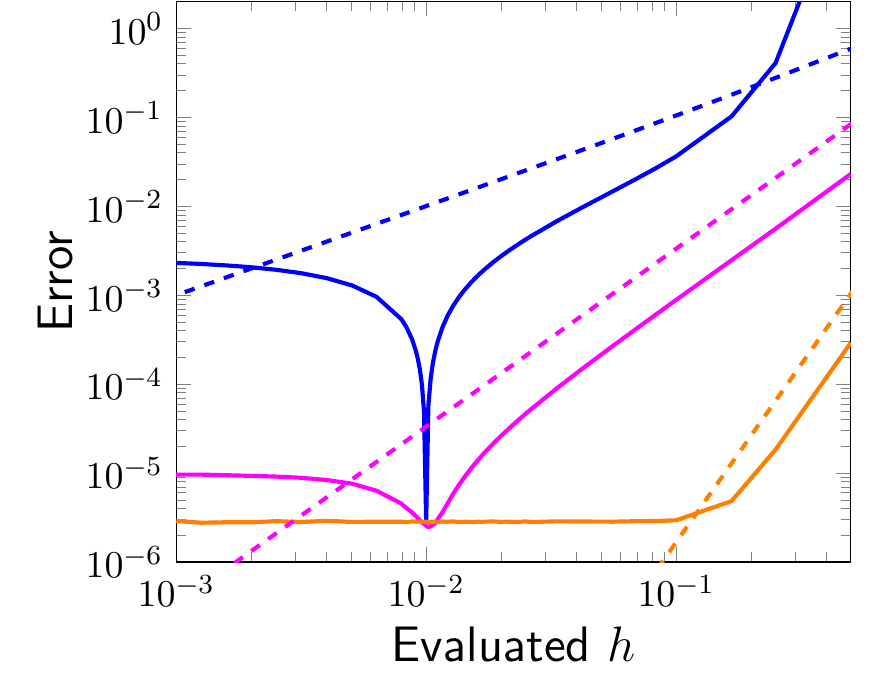}
    }} 
    \subfloat[Trained at $\Delta t$ = 0.2] {{ \includegraphics[width=0.315\linewidth]{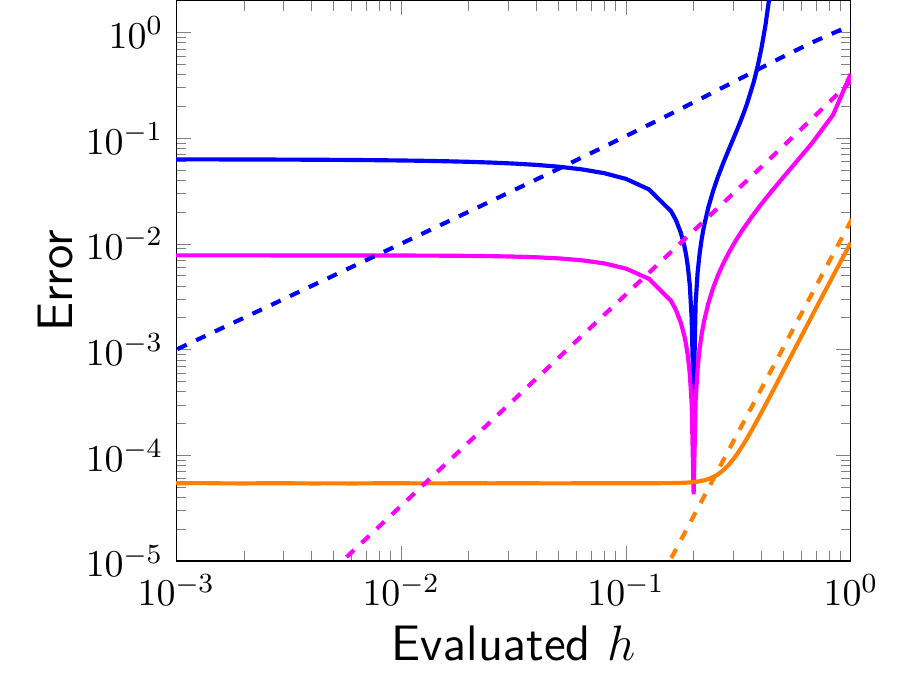}
    }} 
    \subfloat[Trained at $\Delta t = 0.5$] {{  \includegraphics[width=0.3\linewidth]{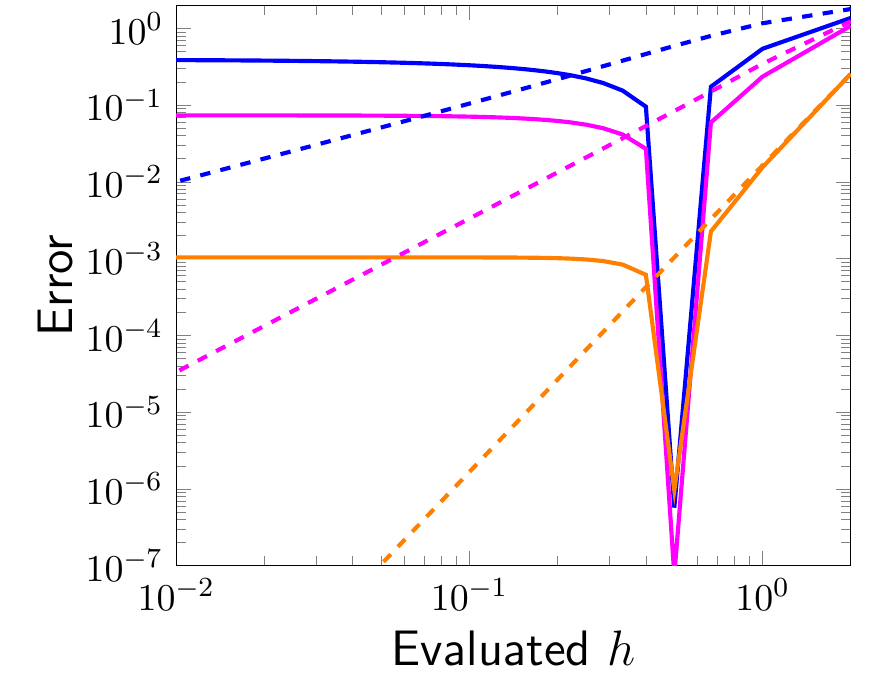}
    }} 
    \caption{\textbf{Supplementary Fig. 1: Harmonic oscillator, additional convergence tests.} We demonstrate the convergence test on the harmonic oscillator, with additional examples where the ODE-Net models are now trained on data spaced apart by different $\Delta t$. During evaluation, depending on the $\Delta t$ in the data, different ODE-Nets overfit to the data. At $h = \Delta t = 0.5$, the RK4-Net also overfits to the training~data.
    }
    \label{fig:harmonic_oscillator_additional convergence}
\end{figure}

\begin{figure}[!ht]
    \centering
    \includegraphics[scale=0.7]{cont_tikz_figs/tikz_figs/pendulum_2dof_inference_legend_3integrators.pdf} 
    \subfloat[Trained $\Delta t$ = 0.01] {{ \includegraphics[width=0.3\linewidth]{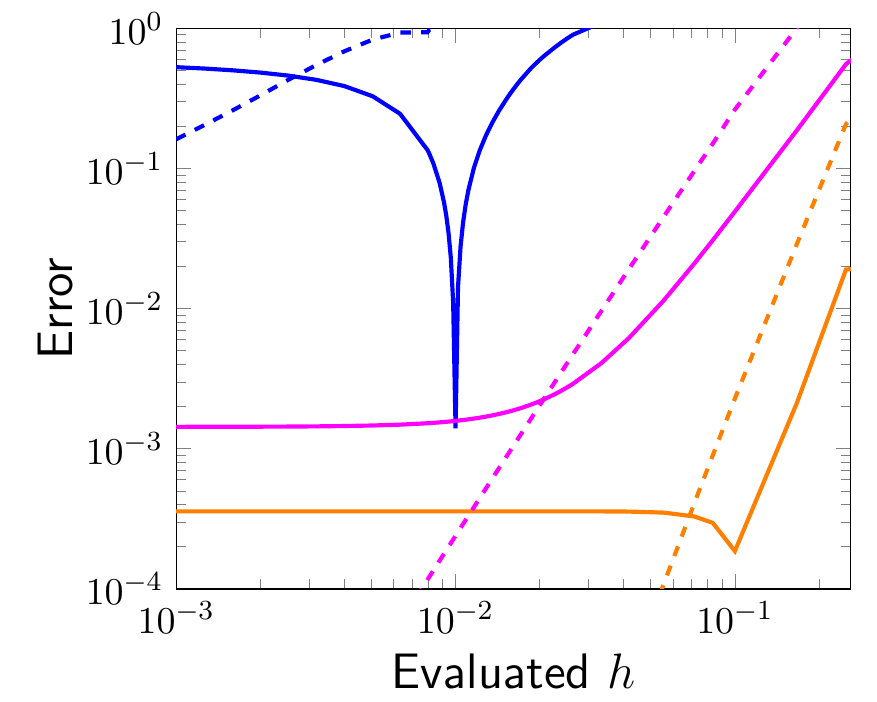}
    }} 
    \subfloat[Trained $\Delta t$ = 0.05] {{ \includegraphics[width=0.3\linewidth]{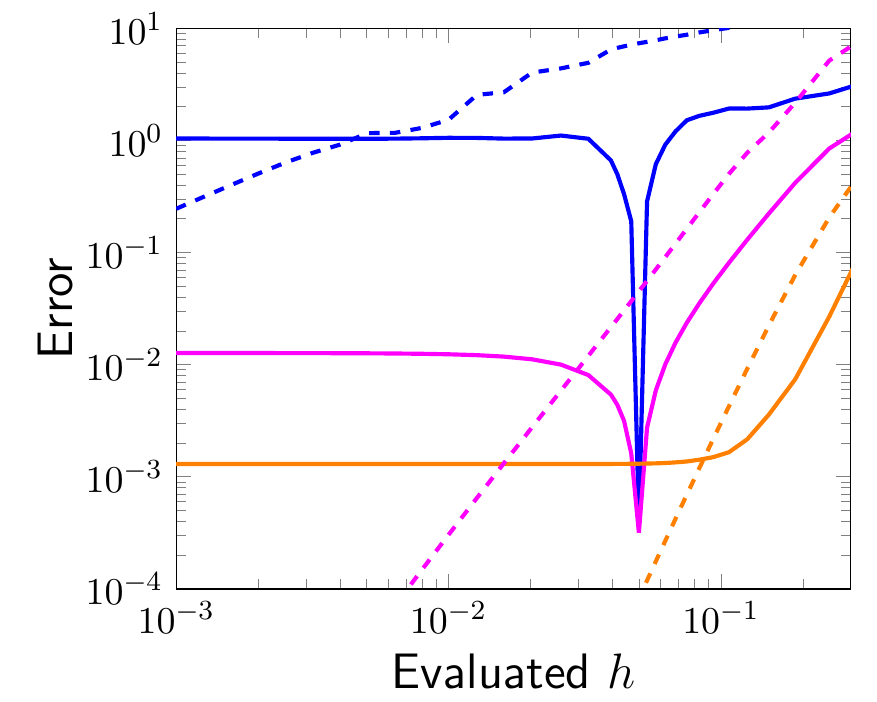}
    }} 
    \subfloat[Trained at $\Delta t = 0.5$] {{  \includegraphics[width=0.3\linewidth]{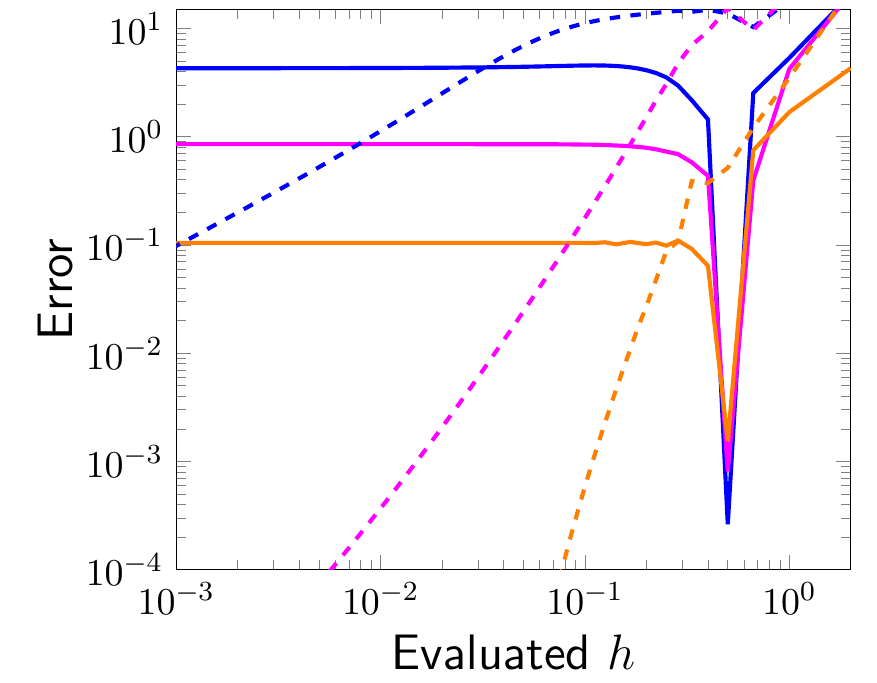}
    }} 
    \caption{\textbf{Supplementary Fig. 2: Non-linear pendulum, additional convergence tests.} We demonstrate the convergence test on the non-linear pendulum, with additional examples where the ODE-Net models are now trained on data spaced apart by different $\Delta t$. Depending on the $\Delta t$ in the data, different ODE-Nets overfit to the data. At $h = \Delta t = 0.5$, the RK4-Net also overfits to the training~data.
    }
    \label{fig:nonlinear_pendulum_additional convergence}
\end{figure}

\begin{figure}[!ht]
    \centering
    \includegraphics[scale=0.7]{cont_tikz_figs/tikz_figs/pendulum_2dof_inference_legend_3integrators.pdf} 
    \subfloat[Trained $\Delta t$ = 0.01] {{ \includegraphics[width=0.3\linewidth]{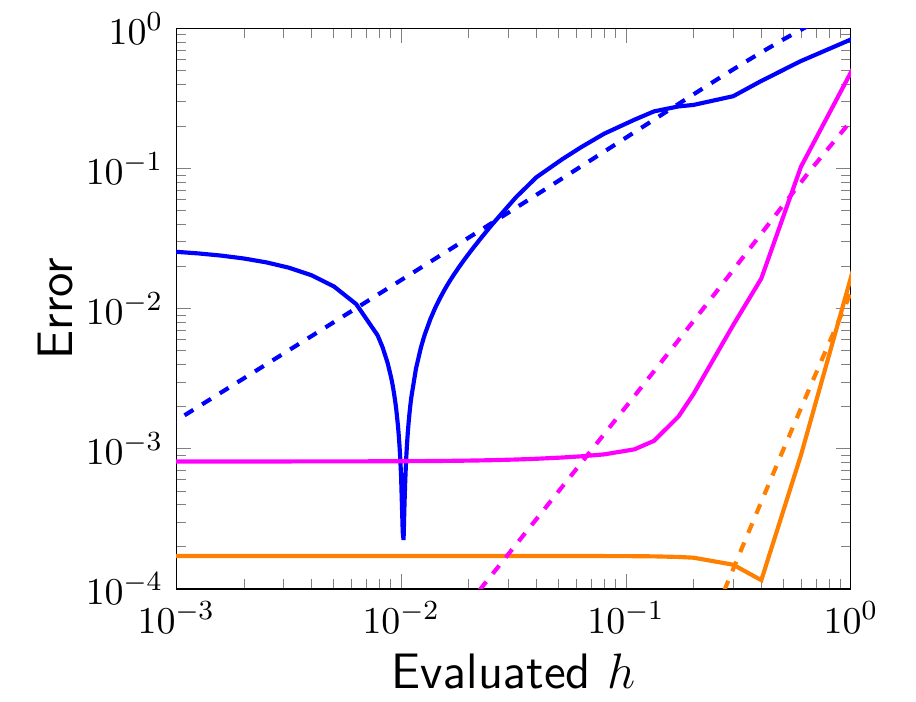}
    }} 
    \subfloat[Trained at $\Delta t$ = 0.05] {{ \includegraphics[width=0.3\linewidth]{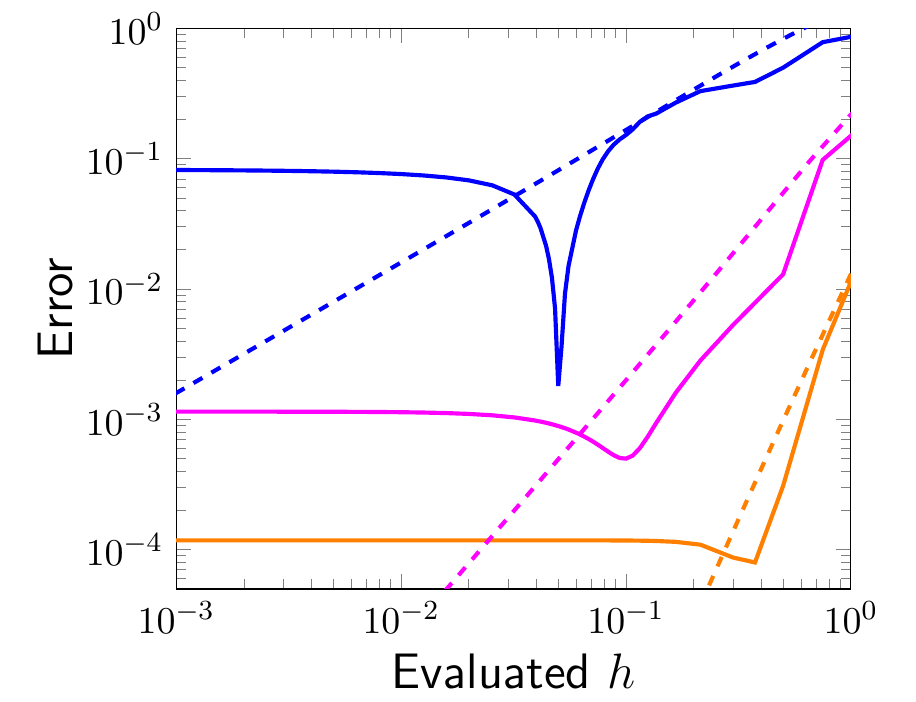}
    }} 
    \subfloat[Trained at $\Delta t$ = 0.2] {{ \includegraphics[width=0.3\linewidth]{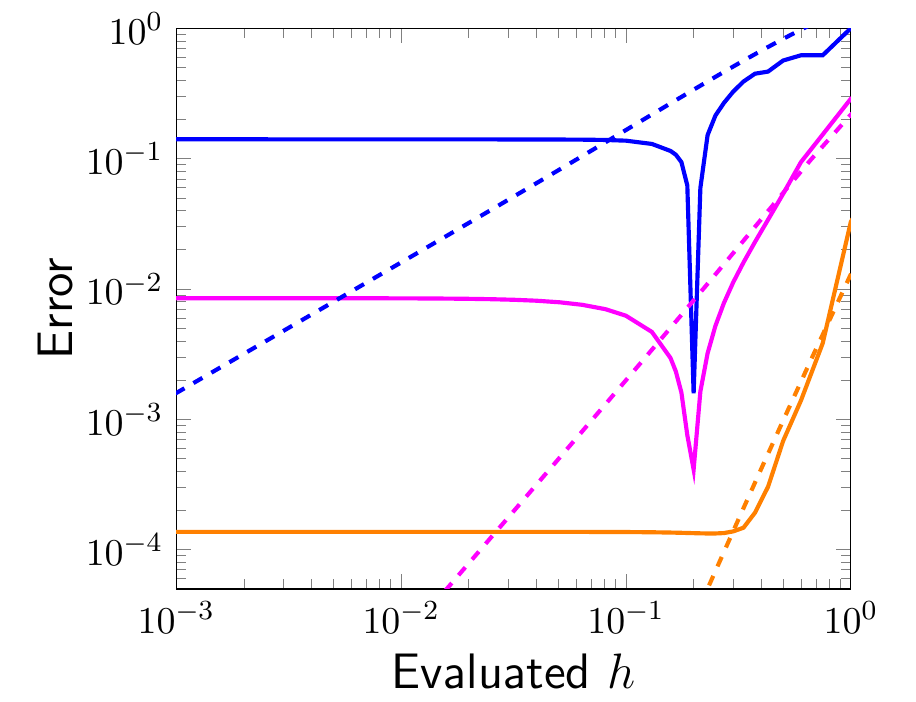}
    }} 
    \caption{\textbf{Supplementary Fig. 3: Lotka-Volterra, additional convergence tests.} We demonstrate the convergence test on the Lotka-Volterra equations, with additional examples where the ODE-Net models are now trained on data spaced apart by different $\Delta t$. While Euler-Net overfits to the training data even when the $\Delta t$ in the training data is small, Midpoint-Net is able to learn a continuous model for higher~$\Delta t$.
    }
    \label{fig:lv_additional convergence}
\end{figure}

\begin{figure}[!ht]
    \centering
    \includegraphics[scale=0.7]{cont_tikz_figs/tikz_figs/pendulum_2dof_inference_legend_3integrators.pdf} 
    \subfloat[Trained $\Delta t$ = 0.001] {{ \includegraphics[width=0.3\linewidth]{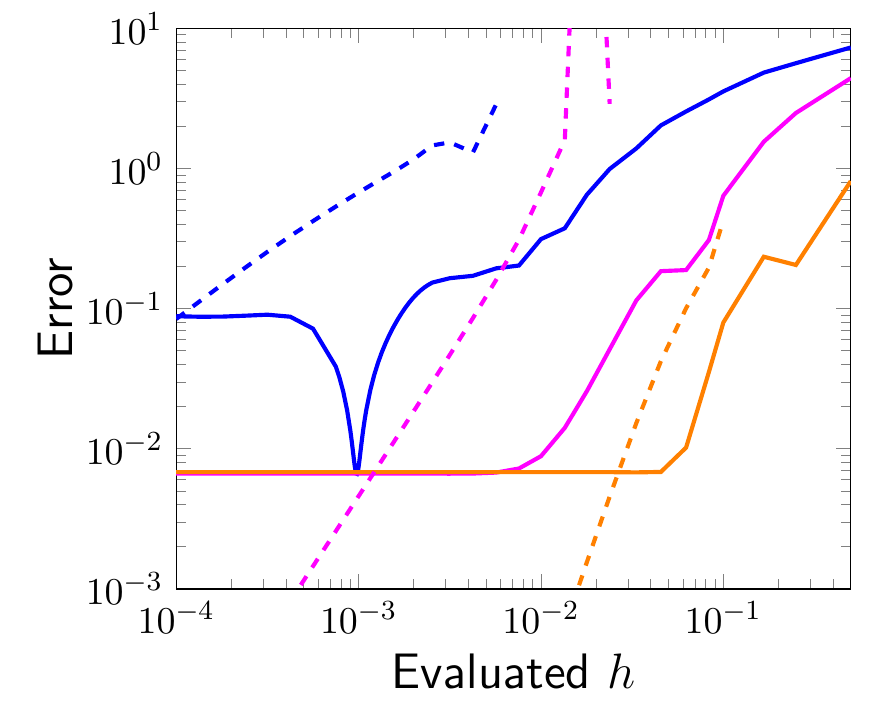}
    }} 
    \subfloat[Trained $\Delta t$ = 0.01] {{ \includegraphics[width=0.3\linewidth]{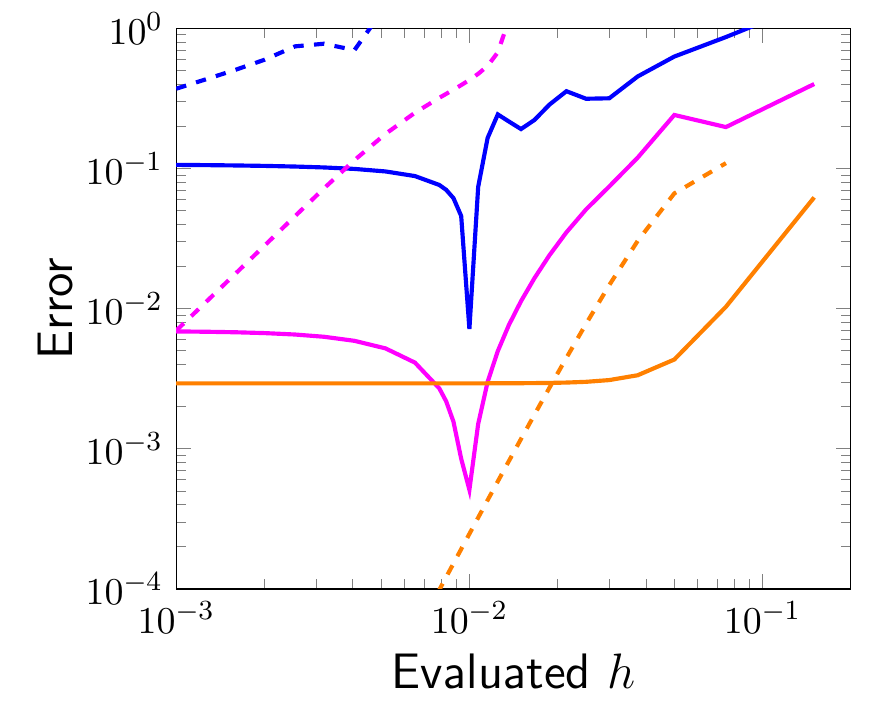}
    }} 
    \subfloat[Trained $\Delta t$ = 0.1] {{ \includegraphics[width=0.3\linewidth]{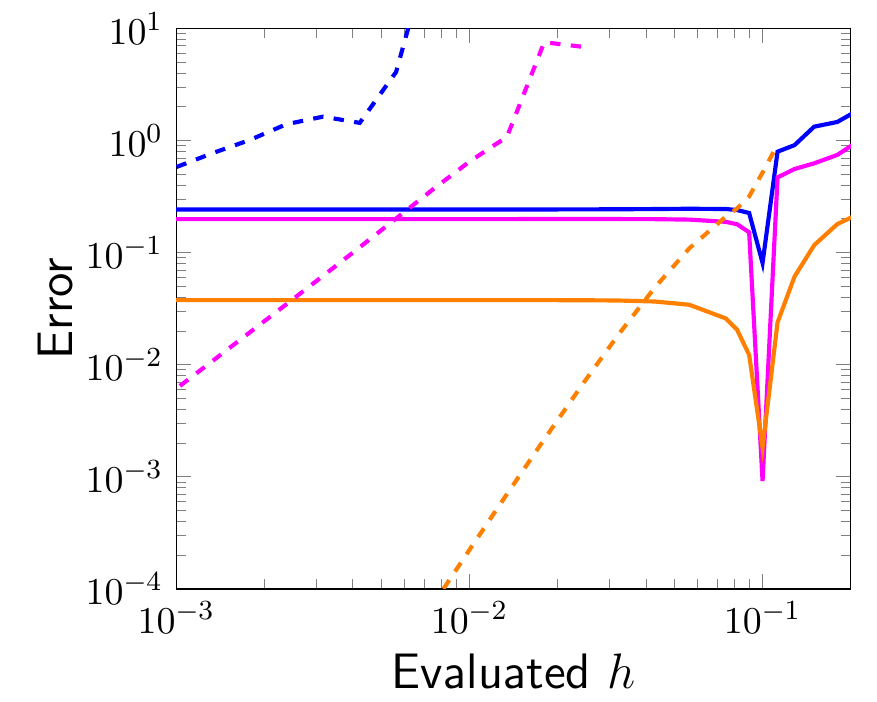}
    }} 
    \caption{\textbf{Supplementary Fig. 4: Cartesian pendulum, additional convergence tests.} We demonstrate the convergence test on the Cartesian pendulum, with additional examples where the ODE-Net models are now trained on data spaced apart by different $\Delta t$. This is a \textit{stiff} system which can be challenging to solve with standard numerical integration schemes. Note that the NN models (Midpoint-Net and RK4-Net) have lower error than the baseline numerical integrators until the error saturates and they converge to a fixed~value.
    }
    \label{fig:cartesian_pendulum_additional convergence}
\end{figure}

In this section, we include additional examples of convergence tests for the different systems studied in~\nameref{subsec:additional_systems}. 
Here, the ODE-Net models are trained on training data spaced apart by different $\Delta t$ for each system; and we also include examples using the (explicit) Midpoint numerical scheme, a second-order integration scheme.
In the Midpoint scheme, the next timestep is approximated as

\begin{equation}
x_{t+1} = x_t + h \mathcal{N}\left(x_t + \frac{h}{2}, x_t + \frac{h}{2} \mathcal{N}(x_t; \theta); \theta\right).
\end{equation}
We observe that, when evaluated at different $h$, different numerical integrator ODE-Nets either overfit to the discrete data points, or learn meaningfully continuous dynamics. 
We also observe that, while moving to higher-order ODE-Net integrators can add inductive biases to better learn continuous dynamics, these integrators (such as RK4) can still overfit to the temporal discretization, when the $\Delta t$ between training data points is large enough. 
In each figure, we also include comparisons to the baseline numerical~integrator.

In~\fref{fig:harmonic_oscillator_additional convergence}, we show additional convergence test examples of the harmonic oscillator trained at different $\Delta t$. When $\Delta t = 0.01$, Euler-Net fails the convergence test (low error at $ h = \Delta t$, but high everywhere else) while Midpoint-Net and RK4-Net pass the convergence test (error monotonically converges to a fixed value). Once $\Delta t$ increases to $\Delta t = 0.2$, Midpoint-Net now also fails the convergence test, while RK4-Net passes. Finally, once $\Delta t$ increases to $\Delta t = 0.5$, RK4-Net also fails the convergence test. 

In~\fref{fig:nonlinear_pendulum_additional convergence}, we show additional convergence test examples of the non-linear pendulum trained at different $\Delta t$. We observe a similar pattern as the harmonic oscillator: at $\Delta t = 0.01$, only the Euler-Net fails the convergence test while Midpoint-Net and RK4-Net pass. As $\Delta t$ increases to 0.05, Midpoint-Net also fails the convergence test. Finally, once again, as $\Delta t = 0.5$, RK4-Net also fails the convergence test.

In~\fref{fig:lv_additional convergence}, we show additional convergence test examples of the Lotka-Volterra system trained at different $\Delta t$. Similar to the other cases, at low $\Delta t$, only Euler-Net fails the convergence test (while Midpoint-Net and RK4-Net pass). As $\Delta t$ increases, Midpoint-Net also eventually fails the convergence test.

In~\fref{fig:cartesian_pendulum_additional convergence}, we show additional convergence test examples of the Cartesian pendulum trained at different $\Delta t$. The Cartesian pendulum is an example of a \textit{stiff} system, so the baseline Euler numerical integrator only starts to converge when the evaluated $h$ is small. Here, we see a similar pattern: at very low $\Delta t = 0.001$, Euler-Net fails the convergence test while Midpoint-Net and RK4-Net pass the test. Once $\Delta t = 0.1$, Midpoint-Net also fails the convergence test. Note that the ODE-Nets, when they pass the convergence test, all have lower error than the numerical integrator for numerous evaluated $h$. Since the Cartesian pendulum is a stiff system, the ODE-Nets can fail the convergence test at smaller $\Delta t$ than the other systems of study.

\section*{Supplementary Note 4: Algorithm for the Convergence Test}
\label{sxn:app_algorithms}

Algorithm~\ref{alg:forward} describes our main algorithm, discussed in the main text. 
$ErrorMetric(x_{val}, \bar{x})$ is an abstract function that compares the inferred trajectory against the test trajectory. The function $ErrorMetric.Nudge(h)$ finds the closest timestep size that will evenly divide the sampled points of the test trajectory. For example, if the error metric is using every 9th points, the Nudge function makes sure that $h$ divides evenly into $9\Delta t$.
Algorithm~2 describes how to integrate the convergence test into a simple model discovery loop to automatically find an ODE-Net which is continuous. This algorithm assumes a routine called $Train()$ whose inputs are training data and the desired $Order$ of accuracy of the time integration. For ODE-Nets, $Order$ corresponds to selecting the order of the Runge-Kutta scheme. For SINDy, $Order$ corresponds to selecting the finite difference approximation.

\begin{figure}[!ht]
	\begin{center}
 		\scalebox{0.75}{
			\begin{algorithm}[H]
				\SetAlgoLined
				\DontPrintSemicolon
				\KwData{$F$: Model \tcp*[r]{The neural network}\\
                    $\mathtt{scheme}$: Function \tcp*[r]{Which integration method, e.g., ``RK4''}\\
					$T$: float  \tcp*{Duration of validation trajectory}
					$x(0)$: Array \tcp*[r]{Validation trajectory start point}\\
					$x_{val}$: Array \tcp*[r]{Validation trajectory}\\
					$\Delta t$: float \tcp*[r]{Training timestep} \\
					$m$: int  \tcp*[r]{Span of $h$s to test} \\
                    $\epsilon$: float \tcp*[r]{Heuristic threshold on error margin}
				}
				\Begin{
					
					$\mathtt{h\_test\_array} = \{1.1^i \Delta t \,\,|\,\, i \in \left(-m...0...m\right)\}$  \tcp*[r]{Logarithmic spacing around $\Delta t$}
					\ForEach{$h \in \mathtt{h\_test\_array}$}{
						$h =  ErrowMetric.Nudge(h)$  \tcp*[r]{Tweak to line up sampled points}
						$\bar{x} = x(0)$\;
						\For{$j=0; j < N; j\mathrel{++}$}{
							$\bar{x} = \mathtt{scheme}(F,\bar{x},h)$ \tcp*[r]{Integrate by iterating}
						}
						$error[h] = ErrorMetric(x_{val}, \bar{x})$ \tcp*[r]{Error against validation datapoints}
					}
					Plot $(log(h), log(error))$  \tcp*[r]{Optional convergence plot}
					\uIf{$error[h]\leq (1+\epsilon)\,error[\Delta t] \,\,\, \forall \,h<\Delta t$}{
						return Pass   \tcp*[r]{Assert error for smaller $h$}
					}
					\Else{
						return Fail\;
					}
					\caption{\label{alg:forward} Convergence test.}
				}
		\end{algorithm}
 		}
	\end{center}
\end{figure}

\begin{figure}[!ht]
	\begin{center}
 		\scalebox{0.75}{
			\begin{algorithm}[H]
				\SetAlgoLined
				\DontPrintSemicolon
				\KwData{
					$x_{train}$: Array  \tcp*{Training data} \\
					$x_{val}$: Array \tcp*[r]{Validation data} \\
					$\Delta t$: float \tcp*[r]{Timestep in data}
				}
				\Begin{
				    Order = 1 \;
					\While{ Order $<$ QuitOrder}{
						Model = Train($x_{train}$, Order) \tcp*[r]{Train model given temporal acc.~order}
						Converged = RunConvergenceTest(Model, $x_{val}$, $\Delta t$) \tcp*[r]{Test model}
						\uIf{Converged}{
    						return Model\;
    					}
    					\Else{
    						Order += 1 \tcp*[r]{Try again with higher temporal accuracy.}
    					}
					}
					return Model, Error("Failed to converge before temporal accuracy order ", QuitOrder) \;
					\caption{\label{alg:forward2} Continuity testing within a model discovery loop.
					}
				}
		\end{algorithm}
 		}
	\end{center}
\end{figure}

\paragraph{Choice of error metrics.}
One can use different metrics for error estimation for consistency/ convergence. 
Possible options are:
\begin{align}
\text{End point error} & = \left\| x(t_N) - \bar{x}_N \right\|_2 \\
\text{Max error across data points} & = \max_{1,2,\dots,N} \left\| x(t_n) - \bar{x}_n \right\|_2 \\
\text{Sum of errors across data points} & = \sum_{1,2,\dots,N} \left\| x(t_n) - \bar{x}_n \right\|_2^2  .
\end{align}
It is possible to run the convergence test with any norm. 
Our default option is to the $\ell_2$ norm. 
The first metric is easy to implement. 
However, in the presence of noisy data, the last metric is preferable, because it will average out the impact of noise on individual data points.
This requires adjusting the integration algorithm to nudge timesteps to snap to $t_{n}$ to get the predictions at the necessary times (which is a standard feature in Initial Value Problem, IVP, solvers). 
Taking only a subset of the validation trajectory, such as every fifth data point, is necessary to make the snapping-behavior still allow for variability in $h$.
(Inference timesteps larger than the time gap between data points, $h>\Delta t$, would skip over data points, and the snapping requirement forces an effective timestep size of $\Delta t$.)

\paragraph{Multiple Trajectories.}
Multiple trajectories can be used in training, testing, and validation. As discussed in \nameref{subsec:other_initial_conditions}, divvying the hold-back sets across trajectories is an important technique for verifying generalizibility. The convergence test can be run on each trajectory in the validation set separately, and the combined results can be used to make the decision. This is implemented by an outer loop over Algorithm~1 that repeats the test and evaluates the condition for each tuple of initial condition, duration, and final value. If the convergence test passes for all trajectories in validation set, then the learned model satisfies the convergence test. Otherwise, Algorithm~2 continues by increasing the temporal discretization accuracy and learns a new model again.

It is reasonable to generate a plot by generating the average error across all trajectories for each inference timestep $h$. Note that the  scale of the error is trajectory-dependent. Thus, it is important to evaluate the condition on each trajectory separately.
The value of the error, $b$, as $h\rightarrow 0$, will be different for different trajectories. Thus, when diagnosing problems where a model is not passing the convergence test for certain validation trajectories, it can be useful to generate plots for individual trajectories.

{ \paragraph{Consistency and stability.}  The three critera of classical numerical analysis are consistency, stability, and convergence. In this work, we only introduce a direct test for convergence. We briefly also comment on consistency and stability.

\begin{enumerate}
    \item Consistency, a criterion on one-step error, cannot be \emph{directly} measured in the ML context, because data points cannot be measured at arbitrary small intervals in nonsynthetic examples. Further, we do not expect the error to be 0; the residual error from ML approximation will still exist in the limit.
    Instead, our convergence test \emph{indirectly} measures a similar concept to consistency by implying that that the one-step error does not increase in the limit $h\rightarrow 0$.
    \item Stability of the ML system must be measured and verified separately. By taking long sequences, the convergence test will demonstrate that the ML model can feed onto itself under perturbation of timestep size without blowing up. Further analysis may be desirable for certain applications, such as noise perturbation~\cite{lim2021noisy}. However, stability tests will not prove convergence of an ML model.
\end{enumerate}

}

\section*{Supplementary Note 5: Theoretical Derivations}
\label{sec:theoretical_discussion}

We discuss our theoretical derivations from \nameref{subsec:theory_error_analysis_idealized} for the error bounds for learning in an idealized setting.
The problem domain is simplified to linear ODEs with real scalar values, $\mathrm{d}x/\mathrm{d}t=\lambda x$ and linear models, $\mathcal{N}(x)= wx$. 
The training procedure minimizes the equation,
\begin{equation}
\label{eq:analytical_optimization}
    \min_{w} \sum_{x\in \mathcal{D}} \left(x_{n+1}-
    \text{ODESolve}\left[ wx, \,\,x_0=x_n,\,\, \text{step}=\Delta t,\,\, \texttt{scheme} \right]\right)^2.
\end{equation}
For simplicity, we assume that the dataset has no error and was acquired from the analytical solution, $x(t+\Delta t) = e^{\lambda \Delta t}x(t)$.
Firstly, we demonstrate that optimization of the above equation will yield $w\neq\lambda$.
Secondly, we demonstrate that is possible to estimate the error arising from the time discretization in training by using only the validation dataset.

\subsection*{Supplementary Note 6: Optimization of $w$ does not learn $\lambda$}
\label{sec:appendix_proof_1}



\begin{proof*}[Proof of Lemma \ref{thm:polynomial_root}]
The optimization problem solves the minimization problem,
\begin{align}
w & = \arg \min_{w'} \sum_{x\in \mathcal{D}}\left( e^{\lambda \Delta t}x_n - \mathtt{scheme}(w'x,\Delta t)(x_n)\right)^2 ,
\end{align}
where $w'$ is the variable of optimization, and $\mathtt{scheme}$ is the application of one step of the numerical integrator.
There exists a ``perfect'' model $\mathrm{d}x/\mathrm{d}t=wx$ that minimizes this equation. A realized model $\tilde{w}$ could also have additional residual error, $\tilde{w}=w+\varepsilon$, owing to numerical error, noise in the data, etc. We distinguish this source of error from the error originating from a $w$ that solves the above minimization problem.

For a linear ODE, it is possible in certain configurations to obtain a $w$ that sets this residual to zero for all points $x_n$ in the dataset:
\begin{equation}
    \exists \mathtt{scheme}, \Delta t, w \quad s.t. \quad 0 = e^{\lambda \Delta t} x_n - \mathtt{scheme}(wx,\Delta t)(x_n) \quad \forall x_n .
\end{equation}
Let $p$ denote the order of accuracy of the scheme used for training. A consistent numerical integration scheme will expand a linear ODE into the truncated Taylor series polynomial with $p$ terms. Then, the residual equation is, 
\begin{equation}
    0 = e^{\lambda \Delta t}x_n - \left(\sum_{i=0}^p \frac{\Delta t^i}{i!} w^i \right)x_n \quad \forall x_n .
\end{equation}
For example, Euler, $p=1$, yields a linear equation; for RK2, $p=2$, yields a quadratic equation; and for RK4, $p=4$, the equation is a quartic equation. It is thus apparent that it is possible to obtain a $w$ that solves the equation,
\begin{equation}
\label{eq:polynomial_solution_appx}
    \sum_{i=0}^p \frac{\Delta t^i}{i!} w^i = e^{\lambda \Delta t} ,
\end{equation}
given certain $p$ and $\Delta t$, for any value of $\lambda$.
For odd $p$, there is always such a root. For any $p$, this equation has the solution $w = \lambda$ in the limit $\Delta t\rightarrow 0$. For $p=2$ and $p=4$, the analytical solutions of the roots yield the following conditions for there to be such a root.
For $p=2$, the quadratic formula discriminant condition is $1-e^{\Delta t \lambda}>0$, yielding $\Delta t \leq \log(2)/|\lambda|$. For $p=4$, the discriminant of the quartic solution yields the condition $\Delta t <1.307/|\lambda|$ (where we approximated the rational expression for brevity).

\hfill$\blacksquare$
\end{proof*}
\noindent
Equation~\eqref{eq:polynomial_solution_appx} can be solved analytically up to $p=4$ (see the following); and, beyond that, it can be solved with numerical root finding. It is thus apparent that $w$ is \emph{not equal to} $\lambda$ in general, even with error-free optimization in the absence of data noise. That is, $\lambda$ does not minimize the training loss in the general case.

Note that the solution in Eq.~\eqref{eq:polynomial_solution_appx} can also be adapted for linear ODEs on other domains. For example, for scalar ODEs with complex valued parameters, there is always a solution for $w$, as complex polynomials always have complex roots. Additionally, for any dimension of vector-valued $x$ with matrix $w$ and $\lambda$, there will also always be a solution for $p=1$ (Euler). The following results continue to assume real scalar parameters for simplicity of presentation.

\subsection*{Supplementary Note 7: Example evaluations of the analytical equations}
\label{sec:appendix_example_equations}

We demonstrate error bounds using two different numerical integrators, Forward Euler and RK2 (explicit midpoint).

\paragraph{Example: Forward Euler.}
For Forward Euler, $p=1$. Then, the equation above is simple to solve analytically,
\begin{equation}
w_{euler} = \frac{1}{\Delta t}(1-e^{\lambda \Delta t}), 
\end{equation}
yielding the error between learned parameter and ODE parameter,
\begin{equation}
|w_{euler} - \lambda | = \left|\frac{\Delta t}{2}\lambda^2 +
\frac{\Delta t^2}{6}\lambda^3 + ...\right| .
\end{equation}
The global convergence error at inference time for any test trajectory for any $h$ is,
\begin{equation}
    \mathrm{Error}_{euler}(h) = k\left|\frac{e^{\lambda h} - 1}{h} - \left(\frac{1-e^{\lambda \Delta t}}{\Delta t} + \varepsilon\right)\right| .
\end{equation}
When $h=\Delta t$, we see that $\mathrm{Error}_{euler}(\Delta t)=|\varepsilon|$.
The value for $b$ is,
\begin{equation}
    b_{euler} = k \left| \frac{1}{\Delta t}(1-e^{\lambda \Delta t}) + \varepsilon - \lambda\right| + \mathcal{O}(h^{q}) .
\end{equation}
Taking a series expansion in terms of $\Delta t$,
\begin{equation}
    b_{euler} = k \left|\varepsilon + \frac{\Delta t\lambda^{2}}{2} + \frac{\Delta t^{2} \lambda^{3}}{6} + O\left(\Delta t^{3}\right)\right| + \mathcal{O}(h^{q}) .
\end{equation}
We can summarize that term as,
\begin{equation}
    b_{euler} = k |\varepsilon| + k \frac{|\lambda|^2}{2} \Delta t + O\left(\Delta t^{2}\right) + \mathcal{O}(h^{q}) .
\end{equation}
Therefore, training using a forward Euler integrator has a convergence error that is linear with respect to both $\Delta t$ and $\varepsilon$. Comparing the difference between $\mathrm{Error}(\Delta t)$ and $b$ allows us to estimate empirically that $|w-\lambda| = \mathcal{O}(\Delta t)$.


\paragraph{Example: RK2 (Explicit Midpoint).}
For RK2, $p=2$, and applying the recursive rule gives us a quadratic polynomial which can be solved analytically,
\begin{equation}
w_{rk2} = \frac{\sqrt{2 e^{\Delta t \lambda} - 1} - 1}{\Delta t} .
\end{equation}
This equation has real roots when $\Delta t \leq \log(2)/|\lambda|$.
Note that there can be two roots; for higher order numerical integrators, there may be multiple parameters that minimize the ML loss function perfectly. The difference between this value and the underling $\lambda$ can be expanded into a series as,
\begin{equation}
|w_{rk2} - \lambda | = \left|\frac{\Delta t^2\lambda^3}{6} - \frac{\Delta t^3\lambda^4}{8} + \mathcal{O}(\Delta t^4)\right| .
\end{equation}
Plugging back into $\mathrm{Error}(h)$, we obtain,
\begin{equation}
    \mathrm{Error}_{rk2}(h) = k\left| w_{rk2}+\varepsilon + h (w_{rk2}+\varepsilon)^2 + \frac{1-e^{\lambda h}}{h} \right| .
\end{equation}
The error at the training timestep can be expressed by a series expansion,
\begin{equation}
    \mathrm{Error}_{rk2}(\Delta t) = k\left| \varepsilon + \Delta t\left( \varepsilon^2/2+\varepsilon\lambda \right) + \mathcal{O}(\Delta t^3) \right| .
\end{equation}
This shows that the error is dominated by the $\varepsilon$ term. When we plug into the limit as $h\rightarrow 0$, we obtain,
\begin{equation}
b_{rk2} = k \left|\varepsilon + \frac{\sqrt{2 e^{\Delta t\lambda} - 1} - 1}{\Delta t} - \lambda\right| + \mathcal{O}(h^q) .
\end{equation}
Series expanding this equation along $\Delta t$ to  summarize, we obtain 
\begin{equation}
b_{rk2} \leq k |\varepsilon| + k \frac{|\lambda|^3}{6}\Delta t^{2} + \mathcal{O}(h^q) .
\end{equation}
Therefore, the observable error due to the numerical discretization for training is $\Delta t^2$. This means that training through RK2 will approximate the continuous operator asymptotically more accurately than Euler.

\section*{Supplementary Note 8: Approximation errors introduced by scalar ODE-Nets}
\label{sec:appendix_proof_2}

Equation~\eqref{eq:polynomial_solution_appx} can  be used to estimate a bound on the error caused by the minimization process.

\begin{proof*}[Proof of Theorem \ref{thm:wminuslambda}]
Increasing the order of the training scheme $p$ decreases the truncation error of the loss by adding more terms to the polynomial equation, Eq.~\eqref{eq:polynomial_solution_appx}. To demonstrate, first recall that the polynomial in Eq.~\eqref{eq:polynomial_solution_appx} is the truncated Taylor series for $e^x$, which we denote by $T_p(x)$. 
Let $R_p(x)$ denote the Taylor remainder term as well, such that $e^x = T_p(x)+R_p(x)$. We can invoke the mean value theorem
for some $0\leq\delta\leq 1$, whose precise value depends on $\lambda\Delta t$, to bound the term,
\begin{equation}
\label{eq:taylor_remainder}
    R_p(\lambda \Delta t) = \frac{e^{\delta \lambda \Delta t} (\lambda \Delta t)^{p+1}}{(p+1)!} .
\end{equation}

We begin by assuming the necessary conditions to obtain a $w$ that exactly satisfies the polynomial equation Eq.~\eqref{eq:polynomial_solution_appx}, where we substitute in $T_p$,
\begin{equation}
\label{eq:polynomial_residual_proof2}
T_p(w\Delta t) - e^{\lambda \Delta t} = 0.
\end{equation}
For arbitrary $w$, we call the value of the left hand side of this equation the polynomial residual.
We can loosely bound the possible roots of $w$ with general theorems of polynomial roots. The coefficients of the polynomial equation are  $a_0=1-e^{\lambda \Delta t}$, $a_1=\Delta t$, and $a_p=\Delta t^p / p!$.  Because $\Delta t>0$, all of the coefficients of polynomial equation are positive, except $a_0$, which may or may not be positive depending on the value of $\lambda$.
There are three situations to consider:
\begin{itemize}
    \item If $\lambda = 0$, then $a_0=0$, and we can trivially solve that $w=0$ is an obvious root. The ML algorithm can exactly solve a constant time series.
    \item If $\lambda > 0$, then $a_0<0$. There is only one negative coefficient, and thus, by Descartes' rule of signs, only one positive root. We assume that our ML optimization algorithm will find the root with the same sign as $\lambda$ if there are multiple roots. In the following, $w>0$ is the only bound we need when $\lambda>0$.
    \item If $\lambda < 0$, then $a_0>0$. Then, all coefficients are positive, and thus, by Descartes' rule of signs, all roots are negative, and $w<0$. We will also require a lower bound on $w$. For odd $p$, we can obtain a lower bound by observing that there is a sign change in the polynomial residual of Eq.~\eqref{eq:polynomial_residual_proof2} between $w=0$ and $w=\lambda$. When $w=0$, the residual is positive,
    \begin{equation}
        T_p(w\Delta=0)-e^{\lambda \Delta t} = 1-e^{\lambda \Delta t} > 0.
    \end{equation}
    When $w=\lambda$, we can use the Taylor remainder term,
    \begin{equation}T_p(\lambda\Delta t)-e^{\lambda \Delta t}=-R(\lambda\Delta t) = -\frac{e^{\delta\lambda\Delta t}\lambda^{p+1}\Delta t^{p+1}}{(p+1)!}.
    \end{equation}
    The sign of this term is opposite the sign of $\lambda^{p+1}$, thus is always negative when $p$ is odd, and positive when $p$ is even. Therefore, for odd $p$ there is a sign flip of the residual between $0$ and $\lambda$, and we can therefore bound $\lambda < w_{odd\,p} <0$. For even $p$, finding such a loose bound is more difficult, as there exist values of $\lambda$ and $\Delta t$ with no solution for $w$. However, we are assuming combinations of $\lambda,\Delta t,$ and $p$ with a solution for $w$. Then, for even $p$, solving for $w$ with $p-1$ and $p+1$ both satisfy $\lambda<w$. By assuming a solution for $w_{even\,p}$, we assume that the error $|w-\lambda|$ solution can be bounded from above by maximum error of the $p-1$ or $p+1$ cases.
    \begin{equation}
    \exists\,\,a\quad s.t. \quad |w_{even\,p}-\lambda| \leq a \max\left(|w_{p-1}-\lambda|,\,|w_{p+1}-\lambda|\right).
    \end{equation}
    That is, if there is a solution for an even $p$, we assume that it is not asymptotically less accurate than a constant bound on $w_{p-1}$ or $w_{p+1}$. We can therefore assume the bound $a\lambda<w_{even\,p}<0$ for a small constant $a>1$. Thus, the bound $a\lambda<w<0$ with a small constant $a\geq 1$ encompasses both even and odd $p$.
\end{itemize}
These give us the following bounds:
$sign(w)=sign(\lambda)$, and $a\lambda<w<0$ when $\lambda<0$.

We will leverage these bounds to make a tighter bound in the following. 

Returning to Eq.~\eqref{eq:polynomial_residual_proof2}, the Taylor expansion allows the following manipulation,
\begin{align}
    T_p(w\Delta t) - e^{\lambda \Delta t} & = 0, \\
    T_p(w\Delta t) - T_p(\lambda \Delta t) - R_p(\lambda \Delta t) & = 0, \\
    T_p(w\Delta t) - T_p(\lambda \Delta t) & = R_p(\lambda \Delta t).
\end{align}
The left hand side can be expanded as so,
\begin{equation}
    T_p(w\Delta t) - T_p(\lambda \Delta t) = \sum_{i=1}^{p} \frac{\Delta t^i}{i!}\left(w^i-\lambda^i\right).
\end{equation}
We can use the factorization $w^i-\lambda^i=(w-\lambda)(w^{i-1}+w^{i-2}\lambda+\dots + \lambda^{i-1})$ to pull out the difference between $w$ and $\lambda$,
\begin{align}
    T_p(w\Delta t) - T_p(\lambda \Delta t) 
    & = \left(\sum_{i=1}^{p} \frac{\Delta t^{i-1}}{i!}\left(\sum_{k=1}^{i}w^{i-k}\lambda^{k-1}\right)\right)\Delta t\left(w-\lambda\right) \\
    & := K(w,\lambda) \Delta t (w-\lambda),
\end{align}
where we define $K(w,\lambda)$ to shorthand the double summation in the factorization.
On the right hand side, we can substitute in the bound of Eq.~\eqref{eq:taylor_remainder}, with $\delta$ between 0 and $1$.
Then, we have the expression for the difference between the $w$ and $\lambda$,
\begin{equation}
w-\lambda = \frac{e^{\delta\lambda\Delta t}\lambda^{p+1}}{(p+1)!K(w,\lambda)}\Delta t^p.
\end{equation}
We see the main term in $\Delta t^p$, but there is still a dependence on $w$ in the denominator. We can now use the above bounds on $w$ as a polynomial root to bound the minimum possible value of $|K(w,\lambda)|$ to produce the upper bound on $|w-\lambda|$. For the two nontrivial ($\lambda\neq 0$) cases:
\begin{enumerate}
    \item In case of $\lambda > 0$, the lower bound of $w>0$ makes $K(w,\lambda)$ the smallest it could be because it is monotonic, yielding the \emph{upper} bound on $w-\lambda$:
    \begin{equation}
        |w-\lambda| < \left|\frac{e^{\delta\lambda\Delta t}\lambda^{p+1}}{(p+1)!K(0,\lambda)}\Delta t^p\right|.
    \end{equation}
    The expression $K(w=0,\lambda)$ with $w=0$ is
    \begin{equation}
        K(0,\lambda)= \left(\sum_{i=1}^{p} \frac{\Delta t^{i-1}}{i!}\lambda^{i-1}\right) .
    \end{equation}
    We can series expand the upper bound in terms of $\Delta t$ to see that this has a leading term in~$\Delta t^p$,
    \begin{equation}
|w-\lambda| \leq \left|\frac{1}{(p+1)!}\left(
    \lambda^{p+1}\Delta t^{p} +
    \left(\delta - \frac{1}{2} \right)\lambda^{p+2}\Delta t^{p+1}
    + \mathcal{O}(\Delta t^{p+2})\right)\right|.
    \end{equation}
    \item In case of $\lambda < 0$, then $w<0$. For $p$ odd, we established that $|w|<|\lambda|$. 
    For even $p$, we assume that there exists a constant $a$ that bounds $w$, if Eq.~\eqref{eq:polynomial_residual_proof2} has a solution for $w$. We thus assume that there exists an $a$ such that $|w| \leq a|\lambda|$ for a small constant $a$. The smallest-magnitude value of $K$ will then occur at some value $w=s\lambda$ for $0<s\leq a$. Then we may bound $K(w,\lambda)$ from the bottom:
\begin{align}
        K(w=s\lambda,\lambda) & = \left(\sum_{i=1}^{p} \frac{\Delta t^{i-1}}{i!}\left(\sum_{k=1}^{i}s^{i-k}\lambda^{i-k}\lambda^{k-1}\right)\right) \\ 
        & = \left(\sum_{i=1}^{p} \frac{\Delta t^{i-1}}{i!}\lambda^{i-1}\left(\sum_{k=1}^{i}s^{i-k}\right)\right) \\
        & = \left(\sum_{i=1}^{p} \frac{\Delta t^{i-1}}{i!}\lambda^{i-1}\left(\frac{s^i-1}{s-1}\right)\right) .
    \end{align}
We can plug this back into expression back into the denominator of Eq.~\eqref{eq:polynomial_solution_appx} to obtain an upper bound for $|w-\lambda|$. By series expanding, we can find that the leading term is the same as a above,
\begin{equation}
|w-\lambda| \leq \left|\frac{1}{(p+1)!}\left(
    \lambda^{p+1}\Delta t^{p} +
    \left(\delta - \frac{1+s}{2} \right)\lambda^{p+2}\Delta t^{p+1}
    + \mathcal{O}(\Delta t^{p+2})\right)\right|.
    \end{equation}
    Of note, neither $\delta$ nor $s$ (nor $a$) are in the leading term of the expansion; any range of possible values yields the same leading term.
\end{enumerate}
Therefore, for both $\lambda>0$ and $\lambda<0$, the leading term on the bound is the same.
As $\delta\in[0,1]$ and $s\in(0,a]$ are both small constants, we can thus bound the error with a positive constant times the leading term.
This allows us to bound the difference between the optima of the ML training problem and the true parameters of the target ODE by,
\begin{equation}
    |w - \lambda | \leq c\Delta t^{p} ,
\end{equation}
where $c$ is a constant related to $\lambda$.
When the optimal $w$ is perturbed by numerical error $\varepsilon$, the error between the encountered $\tilde{w}$ and the underlying ODE parameter is,
\begin{equation}
|\tilde{w}-\lambda | = |w+\varepsilon-\lambda | \leq |w-\lambda| + |\varepsilon| \leq c \Delta t^p + |\varepsilon|.
\end{equation}
\hfill$\blacksquare$
\end{proof*}


\section*{Supplementary Note 9: Observable quantities and the convergence test}
\label{sec:appendix_proof_error}
In practice, we know $w$, and we can measure $\mathrm{Error}(h)$ using the dataset; but we do not know $\lambda$, and thus we do not know $\varepsilon$. The convergence test allows the practitioner to make two key observations. We can use the above results to estimate $|w-\lambda |$ and the impact of the $\Delta t^p$~term:
\begin{align}
    b & = c |\varepsilon| + c \Delta t^p + \mathcal{O}(h^q), \\ 
    \mathrm{Error}(\Delta t) & = c |\varepsilon| + \mathcal{O}(\varepsilon\Delta t) .
\end{align}
Therefore, the comparison between $b$ and $\mathrm{Error}(\Delta t)$ allows for the quantitative comparison for the magnitude of $c \Delta t^p$ term. This motivates Eq.~\eqref{eq:convergent_criterion}.
%

We can follow traditional numerical analysis to analyze global error, with the caveat that $w\neq \lambda$. We bound the global error, the error over the whole trajectory, by the truncation error. The truncation error, $\tau$, is,
\begin{equation}
\tau(h) = \left|e^{\lambda h}x_0 - \mathtt{scheme}(\tilde{w}x,h)(x_0)\right| .
\end{equation}
As the timestep size decreases, more steps are required to reach the final time $T$. The truncation error thus adds up more times. To examine global error, we use the bound that the truncation error adds up linearly with the number of timesteps $N=T/h$, such that $\mathrm{Error}(h) = k N \tau(h)$, where $k_\tau$ is a constant proportional to the Lipschitz continuity constants of $\lambda x$ and $w x$. Given a trajectory that starts at $x_0$ and runs until time $T$, the full error for numerically solving $\mathrm{d}x/\mathrm{d}t=\tilde{w}x$ compared to solving $\mathrm{d}x/\mathrm{d}t=\lambda x$ is,
\begin{equation}
\mathrm{Error}(h) = k_\tau\frac{T}{h}\left|e^{\lambda h}x_0 - \mathtt{scheme}(\tilde{w}x,h)(x_0)\right| .
\end{equation}
$T$ and $|x_0|$ are also arbitrary constants, so relate them as $k = k_\tau T |x_0|$.
Using the truncated Taylor series polynomial expansion of the integrator of order $q$, we obtain,
\begin{equation}
\label{eq:error_h}
\mathrm{Error}(h) = \frac{k}{h}\left|e^{\lambda h} - \left(\sum_{i=0}^q \frac{h^i}{i!} \tilde{w}^i \right)\right| .
\end{equation}
 Recall that it is possible to use a different integration scheme at inference time and training time. We track the order of convergence of the integrator used in inference, $q$, separately from the training integrator order, $p$, so as to distinguish the terms.

\begin{proof*}[Proof of Corollary \ref{thm:error_dt}]
When we evaluate the global error using the same numerical integration as at training time, $h=\Delta t$ and $q=p$, the global error is dominated by only the numerical error $\varepsilon$. Firstly, we can substitute in the value of $h=\Delta t$ and $q=p$ above,
\begin{equation}
\mathrm{Error}(\Delta t) = \frac{k}{\Delta t}\left|e^{\lambda \Delta t} - \left(\sum_{i=0}^p \frac{\Delta t^i}{i!} \tilde{w}^i \right)\right| .
\end{equation}
Expanding out $\tilde{w}$, 
\begin{equation}
\mathrm{Error}(\Delta t) = \frac{k}{\Delta t}\left|e^{\lambda \Delta t} - \left(\sum_{i=0}^p \frac{\Delta t^i}{i!} \left(w+\varepsilon\right)^i \right)\right| .
\end{equation}
The summation can be expanded as so,
\begin{align}
\sum_{i=0}^p \frac{\Delta t^i}{i!} \left(w+\varepsilon\right)^i
= & 1
 + \Delta t \left(w+\varepsilon\right)
 + \frac{\Delta t^2}{2} \left(w^2+2w\varepsilon+\varepsilon^2\right) 
 \nonumber \\
& + \frac{\Delta t^3}{6} \left(w^3 + 3w^2\varepsilon + 3w\varepsilon^2 + \varepsilon^3\right)
 + ... \nonumber \\
& + \frac{\Delta t^p}{p!} \left(w^p + pw^{p-1}\varepsilon +\dots +\varepsilon^p\right) .
\end{align}
We can group the terms of $w^i$ to obtain,
\begin{align}
\sum_{i=0}^p \frac{\Delta t^i}{i!} \left(w+\varepsilon\right)^i
= & \left(\sum_{i=0}^p \frac{\Delta t^i}{i!} w^i \right)
 + \Delta t \varepsilon  + \frac{\Delta t^2}{2} \left(2w\varepsilon+\varepsilon^2\right)  \nonumber \\
&
 + \frac{\Delta t^3}{6} \left(3w^2\varepsilon + 3w\varepsilon^2 + \varepsilon^3\right) 
 + ...
 + \frac{\Delta t^p}{p!} \left(pw^{p-1}\varepsilon + \dots +\varepsilon^p\right) .
\end{align}
Thus, we see have the original polynomial in $w$, with additional terms in $\varepsilon$. We truncate the remainder terms of higher powers of $\Delta t$ to simplify to obtain,
\begin{align}
\sum_{i=0}^p \frac{\Delta t^i}{i!} \left(w+\varepsilon\right)^i
= \left(\sum_{i=0}^p \frac{\Delta t^i}{i!} w^i\right) 
 + \Delta t \varepsilon + \mathcal{O}(\Delta t^2\left(w\varepsilon+\varepsilon^2\right)) .
\end{align}
Inserting back into the equation for error, we obtain,
\begin{equation}
\mathrm{Error}(\Delta t) = \frac{k}{\Delta t}\left| \left(e^{\lambda \Delta t} - 
\sum_{i=0}^p \frac{\Delta t^i}{i!} w^i \right)
 - \Delta t \varepsilon + \mathcal{O}(\Delta t^2\left(w\varepsilon+\varepsilon^2\right))
 \right| .
\end{equation}
Recall that $w$ was obtained by the ML process to solve the first term above exactly. Then, by canceling out a factor $\Delta t$, we have the error bound,
\begin{equation}
\mathrm{Error}(\Delta t) %
 = k | \varepsilon | + \mathcal{O}(\Delta t w \varepsilon + \Delta t \varepsilon^2).
\end{equation}
 \hfill$\blacksquare$
\end{proof*}

\noindent
When there is no numerical error, i.e., $\varepsilon=0$, the observed error will be zero, $\mathrm{Error}(\Delta t)=0$, since this is the context in which the model was optimized.

\begin{proof*}[Proof of Corollary \ref{thm:error_h_0}]
To obtain the fixed constant $b$ to which the global error converges, as the timestep goes to zero, we can take the limit of $\mathrm{Error}(h)$ from Eq.~\eqref{eq:error_h}:
\begin{align}
    \lim_{h\rightarrow 0} \mathrm{Error}(h) & = \lim_{h\rightarrow 0} \frac{k}{h}\left|e^{\lambda h} - \left(\sum_{i=0}^q \frac{h^i}{i!} w^i \right)\right|.
\end{align}
As in the previous proof, we can again use the Taylor polynomial $T_q(wh)$ and Taylor remainder term $R_q(wh)$ to easily compare values with $e^{\lambda h}$:
\begin{align}
     \lim_{h\rightarrow 0} \mathrm{Error}(h) & = \lim_{h\rightarrow 0} \frac{k}{h}\left|
     e^{\lambda h} - T_q(wh)\right| \\
     & = \lim_{h\rightarrow 0} \frac{k}{h}\left|
     e^{\lambda h} - e^{w h} + e^{w h}- T_q(wh)\right| \\
     & = \lim_{h\rightarrow 0} k\left|
     \frac{e^{\lambda h} - e^{w h}}{h} + \frac{R_q(wh)}{h}\right|.
\end{align}
We can then invoke Taylor's theorem again to bound the remainder term: there exists a $\delta$ between 0 and $wh$ such that $R_q(wh)=e^\delta (wh)^{q+1}/(q+1)!$. It follows that
\begin{align}
 \lim_{h\rightarrow 0} \mathrm{Error}(h) = \lim_{h\rightarrow 0}  k\left|
     \frac{e^{\lambda h} - e^{w h}}{h} + e^\delta\frac{w^{q+1}}{(q+1)!}h^q \right|.
\end{align}
The term multiplying $h^q$ is a bounded constant. The limit of $(e^{\lambda h}-e^{wh})/h=\lambda-w$.
The limit of the error thus approaches the constant equal to the difference between the learned parameter and ODE parameter at a rate of $h^q$,
\begin{equation}
    \lim_{h\rightarrow 0} \mathrm{Error}(h) = k |w-\lambda| + \mathcal{O}(h^q) .
\end{equation}
(The rate $h^q$ can be observed and measured in the convergence test.)
When there is an additional error $\varepsilon$, the limit is the same but for $\tilde{w}$ appearing in the equation. Then, the error separates into two terms,
\begin{equation}
    \lim_{h\rightarrow 0} \mathrm{Error}(h) = k |\tilde{w}-\lambda| + \mathcal{O}(h^q) \leq k |w-\lambda| + k |\varepsilon| + \mathcal{O}(h^q).
\end{equation}
Substituting the error bounds of Theorem 2, the limit of the global error can also be represented~by
\begin{equation}
    \lim_{h\rightarrow 0} \mathrm{Error}(h) \leq k c \Delta t^p + k |\varepsilon| + \mathcal{O}(h^q).
\end{equation}
 \hfill$\blacksquare$
\end{proof*}

\noindent
In standard numerical analysis, we would have set $w=\lambda$, and we would have proved that our numerical scheme converges to zero at a rate of $q$ for linear ODEs. However, for a learned model, that is not the case. The approached value of $b$ is the same for any $q$, meaning that any numerical integrator used when performing the convergence test will yield similar results in the limit of the approach value.

\color{black}